%% file: main.tex
\documentclass{article}

\usepackage[preprint]{neurips_data_2024}

\usepackage[utf8]{inputenc} 
\usepackage[T1]{fontenc}    
\usepackage[hidelinks]{hyperref}       
\usepackage{url}            
\usepackage{booktabs}       
\usepackage{amsfonts}       
\usepackage{nicefrac}       
\usepackage{microtype}      
\usepackage{xcolor}         
\usepackage{amsmath,amssymb,amsfonts}
\DeclareMathOperator*{\argmin}{arg\,min}
\usepackage{graphicx}
\usepackage{multirow}
\usepackage{algpseudocode}
\usepackage{natbib}
\setcitestyle{numbers,square,comma}
\usepackage{array}
\newcolumntype{C}[1]{>{\centering\arraybackslash}p{#1}}

\usepackage{makecell}
\usepackage{wrapfig}

\usepackage{graphicx}
\usepackage{subcaption}
\usepackage{caption}

\definecolor{lightblue}{HTML}{b3e5fc}
\definecolor{lightyellow}{HTML}{ffeeaa}
\definecolor{lightgreen}{HTML}{c6e9af}

\title{GCondenser: Benchmarking Graph Condensation}

\author{
Yilun Liu, Ruihong Qiu, and Zi Huang\\
\texttt{\{yilun.liu, r.qiu, helen.huang\}@uq.edu.au}\\
The University of Queensland
}

\begin{document}

\maketitle

\input{./sections/abstract.tex}
\input{./sections/introduction.tex}
\input{./sections/preliminary.tex}
\input{./sections/design.tex}
\input{./sections/experiments.tex}
\input{./sections/related_work.tex}
\input{./sections/conclusion_future.tex}

\bibliographystyle{abbrv}
\bibliography{ref}

\newpage
\section*{Checklist}

\begin{enumerate}

\item For all authors...
\begin{enumerate}
  \item Do the main claims made in the abstract and introduction accurately reflect the paper's contributions and scope?
    \answerYes{}
  \item Did you describe the limitations of your work?
    \answerYes{} Section~\ref{sec:conclusion}
  \item Did you discuss any potential negative societal impacts of your work?
    \answerYes{} Section~\ref{sec:conclusion}
  \item Have you read the ethics review guidelines and ensured that your paper conforms to them?
    \answerYes{}
\end{enumerate}

\item If you are including theoretical results...
\begin{enumerate}
  \item Did you state the full set of assumptions of all theoretical results?
    \answerNA{}
	\item Did you include complete proofs of all theoretical results?
    \answerNA{}
\end{enumerate}

\item If you ran experiments (e.g. for benchmarks)...
\begin{enumerate}
  \item Did you include the code, data, and instructions needed to reproduce the main experimental results (either in the supplemental material or as a URL)?
    \answerYes{}
  \item Did you specify all the training details (e.g., data splits, hyperparameters, how they were chosen)?
    \answerYes{}
	\item Did you report error bars (e.g., with respect to the random seed after running experiments multiple times)?
    \answerYes{}
	\item Did you include the total amount of compute and the type of resources used (e.g., type of GPUs, internal cluster, or cloud provider)?
    \answerYes{}
\end{enumerate}

\item If you are using existing assets (e.g., code, data, models) or curating/releasing new assets...
\begin{enumerate}
  \item If your work uses existing assets, did you cite the creators?
    \answerYes{}
  \item Did you mention the license of the assets?
    \answerYes{}
  \item Did you include any new assets either in the supplemental material or as a URL?
    \answerYes{}
  \item Did you discuss whether and how consent was obtained from people whose data you're using/curating?
    \answerYes{}
  \item Did you discuss whether the data you are using/curating contains personally identifiable information or offensive content?
    \answerYes{}
\end{enumerate}

\item If you used crowdsourcing or conducted research with human subjects...
\begin{enumerate}
  \item Did you include the full text of instructions given to participants and screenshots, if applicable?
    \answerNA{}
  \item Did you describe any potential participant risks, with links to Institutional Review Board (IRB) approvals, if applicable?
    \answerNA{}
  \item Did you include the estimated hourly wage paid to participants and the total amount spent on participant compensation?
    \answerNA{}
\end{enumerate}

\end{enumerate}

\newpage
\input{./sections/appendix.tex}


\end{document}

%% file: sections/abstract.tex
\begin{abstract}
Large-scale graphs are valuable for graph representation learning, yet the abundant data in these graphs hinders the efficiency of the training process. Graph condensation (GC) alleviates this issue by compressing the large graph into a significantly smaller one that still supports effective model training. Although recent research has introduced various approaches to improve the effectiveness of the condensed graph, evaluations in a more comprehensive and practical manner are not sufficiently explored. This paper proposes the first large-scale graph condensation benchmark, \texttt{GCondenser}, to holistically evaluate and compare mainstream GC methods. \texttt{GCondenser} includes a standardised GC paradigm with condensation, validation, and evaluation procedures, as well as straightforward extensions to new GC methods and datasets. Furthermore, a comprehensive study of GC methods is conducted, presenting insights into the different dimensions of condensation effectiveness. \texttt{GCondenser} is open-sourced and available at \url{https://github.com/superallen13/GCondenser}.
\end{abstract}

%% file: sections/introduction.tex
\section{Introduction}
In graph representation learning, while large graphs are ideal for optimising graph neural networks (GNNs), they often result in training inefficiencies. Traditional data-centric solutions typically involve sampling large graphs to create representative subgraphs. Although the size of the sampled subgraph becomes significantly smaller, the quality of learned graph representations based on the sampling cannot be guaranteed. Graph condensation (GC) has recently emerged as a graph reduction technique that generates small graphs with synthetic nodes and edges, enabling efficient and effective training of graph models~\cite{jinGraphCondensationGraph2022a, hashemiComprehensiveSurveyGraph2024a}. This technique enhances the efficiency of various graph learning applications, such as graph continual learning~\cite{liuCaTBalancedContinual2023a, liuPUMAEfficientContinual2023}, inference acceleration~\cite{gaoGraphCondensationInductive2023}, federated graph learning~\cite{panFedGKDUnleashingPower2023} and graph neural architecture search~\cite{dingFasterHyperparameterSearch2022}. 

Although numerous GC methods have been proposed and have shown promising performance~\cite{jinGraphCondensationGraph2022a, jinCondensingGraphsOneStep2022a,liuGraphCondensationReceptive2022, liuCaTBalancedContinual2023a, liuPUMAEfficientContinual2023,zhengStructurefreeGraphCondensation2023a}, there is a lack of diverse perspectives of evaluations and comprehensive comparisons among these methods. Therefore, this paper proposes the first large-scale graph condensation benchmark, \texttt{GCondenser}, which aims to assess GC methods across multiple dimensions, including model training efficacy, condensation efficiency, cross-architecture transferability, and performance in continual graph learning. The benchmark covers a thorough collection of gradient matching methods~\cite{jinGraphCondensationGraph2022a, jinCondensingGraphsOneStep2022a}, distribution matching methods~\cite{liuGraphCondensationReceptive2022, liuCaTBalancedContinual2023a, liuPUMAEfficientContinual2023}, and trajectory matching methods~\cite{zhengStructurefreeGraphCondensation2023a}. Moreover, \texttt{GCondenser} standardise the graph condensation paradigm, encompassing condensation, validation and evaluation processes to ensure unified comparisons and simplify the extension to new methods and datasets. Extensive experiments are conducted with \texttt{GCondenser} on various methods and datasets. The contributions of \texttt{GCondenser} can be summarised as follows:
\begin{itemize}
    \item \textbf{A comprehensive benchmark} is developed to evaluate mainstream GC methods from multiple critical perspectives, including model training effectiveness, condensation efficiency, cross-architecture transferability, performance in continual graph learning etc.
    \item \textbf{A standardised graph condensation paradigm} is proposed, from condensation, and validation, to evaluation processes, simplifying the implementation and comparisons of graph condensation methods on various datasets.
    \item \textbf{Extensive experiments} are conducted, which consider not only the hyperparameters of GC methods but also integral components, such as initialisation strategies for condensation, choices of backbone models, validation, downstream tasks etc.
    \item \textbf{To exploit the potential of existing methods}, a thorough tuning of hyperparameters has been explored to improve the quality of condensed graphs generated by existing GC methods.
    \item \textbf{Insightful findings} are provided. (a) Condensed graphs without edges can achieve state-of-the-art quality and benefit from an efficient optimisation process. (b) Initialisation strategies for condensed graphs also influence the final condensation quality. (c) The selection of condensation backbones is sensitive to different categories of methods. (d) Reliable and high-quality condensed graphs depend on suitable validators.
    \item \texttt{GCondenser} is \textbf{open-sourced} to enable a seamless extension of new methods, graph learning applications and datasets.
\end{itemize}

%% file: sections/preliminary.tex
\section{Preliminary}

For a graph learning problem, a graph is denoted as $\mathcal{G}=\{\boldsymbol{A}, \boldsymbol{X}, \boldsymbol{Y}\}$, where the adjacency matrix $\boldsymbol{A} \in \mathbb{R}^{n \times n}$ denotes the graph structure, and $\boldsymbol{X} \in \mathbb{R}^{n \times d}$ is the $d$-dimensional feature matrix for $n$ nodes. 
$\boldsymbol{Y} \in \mathbb{R}^{n}$ represents node labels from a class set $\mathcal{C}$.

Graph condensation aims to synthesise a smaller graph $\mathcal{G}'=\{\boldsymbol{A}', \boldsymbol{X}', \boldsymbol{Y}'\}$ from a larger original graph $\mathcal{G}=\{\boldsymbol{A}, \boldsymbol{X}, \boldsymbol{Y}\}$ while ensuring that the model trained with $\mathcal{G}'$ performs similarly to the model trained with $\mathcal{G}$. This objective can be described as a bi-level optimisation problem:
\begin{equation}
\label{eq:gc-bi}
    \min_{\mathcal{G}'} \mathcal{L}_{\text{task}}(\mathcal{G}; \boldsymbol\theta') \quad\text{s.t.\ } \boldsymbol\theta' = \argmin_{\boldsymbol\theta}\mathcal{L}_{\text{task}} (\mathcal{G}'; \boldsymbol\theta),
\end{equation}
where the lower level optimisation aims to obtain the graph model weight, $\boldsymbol\theta'$, by minimising the task-related loss, $\mathcal{L}_{\text{task}}$, on the condensed graph, $\mathcal{G}'$. With $\boldsymbol\theta'$, the ultimate goal is to obtain the optimal $\mathcal{G}'$ that minimises $\mathcal{L}_{\text{task}}(\mathcal{G}; \boldsymbol\theta')$ at the upper level. However, a direct tackling of this optimisation problem is nontrivial, which leads to a more convenient and effective matching-based paradigm:
\begin{equation}
\label{eq:gc-match}
    \min_{\mathcal{G}'} \mathcal{L}_\text{match}(\mathcal{G}, \mathcal{G}'; \boldsymbol\theta),
\end{equation}
where $\mathcal{L}_{\text{match}}$ is the loss function to measure the distance of key statistics between $\mathcal{G}$ and $\mathcal{G}'$ given model parameter $\boldsymbol\theta$. Different GC methods can be categorised by the choice of statistics:

\textbf{Gradient matching}~\cite{jinCondensingGraphsOneStep2022a, jinGraphCondensationGraph2022a, yangDoesGraphDistillation2023} methods match model learning gradients between the original graph $\nabla_{\boldsymbol\theta}\mathcal{L}_{\text{task}} (\mathcal{G}; \boldsymbol\theta)$ and the condensed graph $\nabla_{\boldsymbol\theta} \mathcal{L}_{\text{task}}(\mathcal{G}'; \boldsymbol\theta)$. The loss $\mathcal{L}_{\text{match}}$ can be defined as:
\begin{equation}
    \mathcal{L}_{\text{match}} = \text{D}(\nabla_{\boldsymbol\theta} \mathcal{L}_{\text{task}} (\boldsymbol{A}, \boldsymbol{X}, \boldsymbol{Y}; \boldsymbol\theta), \nabla_{\boldsymbol\theta} \mathcal{L}_{\text{task}}(\boldsymbol{A}', \boldsymbol{X}', \boldsymbol{Y}'; \boldsymbol\theta)),
\end{equation}
where $\text{D}(\cdot,\cdot)$ is a distance measure of gradients. Generally, the feature matrix $\boldsymbol{X}'$ is optimised directly with $\mathcal{L}_{\text{match}}$. To obtain the synthetic adjacency matrix $\boldsymbol{A}'$, GCond~\cite{jinGraphCondensationGraph2022a} generates $\boldsymbol{A}'$ by mapping $\boldsymbol{X}'$ with learnable MLPs, $\boldsymbol{A}'_{i,j} = \text{Sigmoid}\left([\text{MLP}(x'_i \oplus x'_j) + \text{MLP}(x'_j \oplus x'_i)]/2\right)$, where $\oplus$ represents the concatenate operation. Alternatively, $\boldsymbol{A}'$ can also be treated as the identity matrix $\boldsymbol{I}$ in the condensation process. SGDD~\cite{yangDoesGraphDistillation2023} generates $\boldsymbol{A}'$ using a learnable generator $\text{GEN}_\phi$ from random noise $\mathcal{Z}$ and optimises this generator by reducing the optimal transport distance~\cite{dongCOPTCoordinatedOptimal2020} between the Laplacian Energy Distributions~\cite{tangRethinkingGraphNeural2022} of $\boldsymbol{A}$ and $\boldsymbol{A}'$. A more recent work GDEM~\cite{liuGraphCondensationEigenbasis2023} matches the eigenbasis between the condensed graph and the original graph to manage the structure learning.

\textbf{Distribution matching}~\cite{liuGraphCondensationReceptive2022, liuCaTBalancedContinual2023a, liuPUMAEfficientContinual2023} methods avoid the slow gradient calculation by aligning $\mathcal{G}$ and $\mathcal{G}'$ in the embedding space, corresponding to $\boldsymbol{E}$ for the original graph and $\boldsymbol{E}'$ for the condensed graph:
\begin{equation}
    \mathcal{L}_\text{match} = \text{D}(\boldsymbol{E}, \boldsymbol{E}'),
\end{equation}
where D can be implemented by distance measurement of empirical distribution of embeddings, such as Maximum Mean Discrepancy (MMD) in~\cite{liuCaTBalancedContinual2023a}, $\sum_{c \in \mathcal{C}} \left\| \text{Mean} \left( \boldsymbol{E}_c \right) - \text{Mean} \left(\boldsymbol{E}'_c \right) \right\|^2$, where the subscript $c$ denotes a specific class.

\textbf{Trajectory matching}~\cite{zhengStructurefreeGraphCondensation2023a,geom} methods aim to align the learning dynamics of models trained with $\mathcal{G}'$ and those with $\mathcal{G}$. This alignment is achieved by matching the parameters of models trained on $\mathcal{G}'$ with the optimal parameters of models trained on $\mathcal{G}$. The trajectory matching loss is defined as:
\begin{equation}
    \mathcal{L}_\text{match} = \text{D}(\boldsymbol\theta^*_{t+N}, \boldsymbol\theta_{t+M}'),
\end{equation}
where D measures the difference between the model parameter $\boldsymbol{\theta}_{t+N}^*$ trained with the original graph $\mathcal{G}$ for $N$ steps, and the model parameter $\boldsymbol{\theta}_{t+M}'$ trained with the condensed graph $\mathcal{G}'$ for $M$ steps from the same starting point of model parameter $\boldsymbol{\theta}_t$. Generally, $\boldsymbol{\theta}_{t+N}^*$ and $\boldsymbol{\theta}_t$ can be generated offline.

\textbf{Eigenbasis matching}~\cite{liuGraphCondensationEigenbasis2023} methods align condensed graphs and original graphs in the spectral domain via eigenbasis:
\begin{equation}
    \mathcal{L}_e = \sum_{k=1}^{K} \left\| \boldsymbol{X}^{\top} \boldsymbol{u}_k \boldsymbol{u}_k^{\top} \boldsymbol{X} - \boldsymbol{X}'^{\top} \boldsymbol{u}_k' \boldsymbol{u}_k'^{\top} \boldsymbol{X}' \right\|_F^2,
\end{equation}
where \( \boldsymbol{u}_k \boldsymbol{u}_k^{\top} \) and \( \boldsymbol{u}_k' \boldsymbol{u}_k'^{\top} \) are the subspaces induced by the \( k \)-th eigenvector in the real and synthetic graphs. A discrimination constraint to preserve the category-level information:
\begin{equation}
    \mathcal{L}_d = \sum_{i=1}^{C} \left( 1 - \frac{\boldsymbol{H}_i^{\top} \cdot \boldsymbol{H}'_i}{\|\boldsymbol{H}_i\| \|\boldsymbol{H}'_i\|} \right) + \sum_{\substack{i,j=1 \\ i \neq j}}^{C} \frac{\boldsymbol{H}_i^{\top} \cdot \boldsymbol{H}'_j}{\|\boldsymbol{H}_i\| \|\boldsymbol{H}'_j\|},
\end{equation}
where \( \boldsymbol{H} = \boldsymbol{Y}^{\top} \boldsymbol{A} \boldsymbol{X} \) and \( \boldsymbol{H}' = \boldsymbol{Y}'^{\top} \sum_{k=1}^{K} (1 - \lambda_k) \boldsymbol{u}_k' \boldsymbol{u}_k'^{\top} \). An additional
regularisation is used to constrain the representation space:
\begin{equation}
    \mathcal{L}_o = \left\| \boldsymbol{U}_K'^{\top} \boldsymbol{U}_K' - \boldsymbol{I}_K \right\|_F^2.
\end{equation}
The overall matching loss function of GDEM is formulated as the weighted sum of three regularisation terms:
\begin{equation}
    \mathcal{L}_\text{match} = \alpha\mathcal{L}_e + \beta\mathcal{L}_d + \gamma\mathcal{L}_o.
\end{equation}

%% file: sections/design.tex
\section{Benchmark Design}
\begin{figure}[!t]
    \centering
    \includegraphics[width=\textwidth]{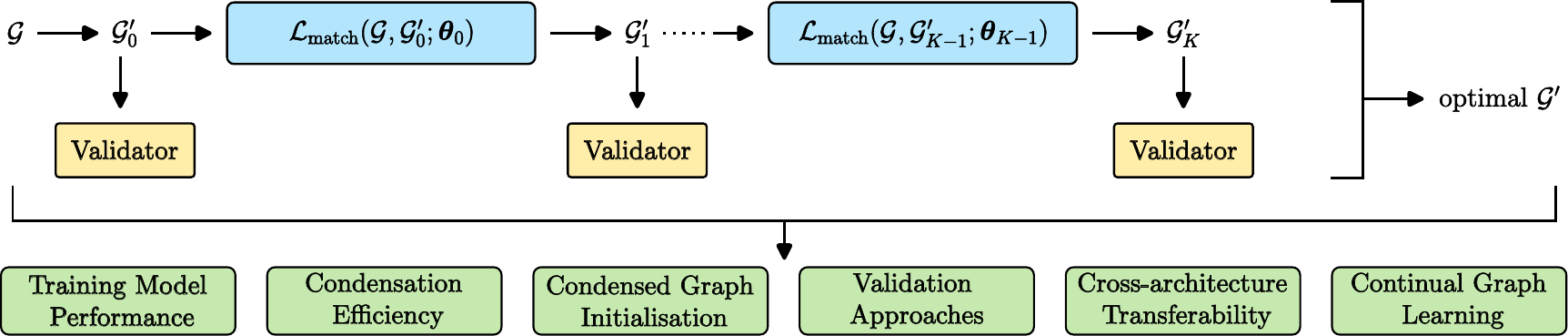}
    \caption{The standardised graph condensation paradigm in \texttt{GCondenser} consists of \colorbox{lightblue}{\strut condensation}, \colorbox{lightyellow}{\strut validation}, and \colorbox{lightgreen}{\strut evaluation} modules. $\mathcal{G}'_{0}$ is first initialised from the original graph $\mathcal{G}$ and matched with the original graph across $K$ initialised model parameter spaces (i.e., from $\boldsymbol{\theta}_0$ to $\boldsymbol{\theta}_{K-1}$). Each matching block represents either multi-step~\cite{jinGraphCondensationGraph2022a} or one-step~\cite{jinCondensingGraphsOneStep2022a} matching.
}
    \label{fig:pipline}
\end{figure}

This section details the benchmark design concerning graph condensation modules and various evaluation methods, as illustrated in Figure~\ref{fig:pipline}. 

\subsection{Graph Condensation Modules}

\textbf{Initialisation of condensed graph}: Most GC methods start from an initialisation of the condensed graph with a given node budget. $\boldsymbol{Y}'$ is typically generated according to a specific label distribution, which could be balanced or proportional to the original graph's label distribution. The initialised node feature $\boldsymbol{X}'$ can be random noise or sampled from $\boldsymbol{X}$ with the given $\boldsymbol{Y}'$. 


\textbf{Condensed graph learning}: During the condensation learning procedure, GC methods rely on a series of randomly initialised graph models to calculate the matching loss. These graph models could be trained using the condensed graph under the multi-step matching strategy.


\textbf{Validation protocol}: 
The validation of graph condensation methods is critical for evaluation. In most existing methods, the validation process is often neglected given that there is no formal validation process. \texttt{GCondenser} develops various validators, such as GNNs and the graph neural tangent kernel (GNTK)~\cite{duGraphNeuralTangent2019a, zhengStructurefreeGraphCondensation2023a}. After updating the condensed graph, a validator is developed using this condensed graph and evaluates the performance of the validation set from the original graph. The condensed graph with the highest validation accuracy is later selected as the optimal condensed graph.

\subsection{Evaluations}

With \texttt{GCondenser}, extensive evaluations are conducted for the following research questions (RQs):

\textbf{RQ1:} How does the performance of models trained on condensed graphs compare to those trained on entire datasets, random graphs, and k-Center graphs across different budget constraints?

\textbf{Training model performance} will be evaluated for existing GC methods. The GNN architecture will not change during condensation, validation, and test phases. In the test phase, a randomly initialised GNN model is trained using the condensed graph and tested on the test set from the original graph. The model’s test accuracy reflects the essential effectiveness of the condensed training graph.

\textbf{RQ2:} What is the condensation efficiency of different graph condensation methods?

\textbf{Condensation efficiency} reflects the required effort for obtaining a high-quality condensed graph. To evaluate the efficiency, the model performance will be compared against the condensation time. The storage efficiency will be directly evident by the model performance against the budget ratio.

\textbf{RQ3:} How do condensed graph initialisation schemes affect the quality of the condensed graph?

\textbf{Different initialisation strategies} decide the starting point of the condensed graph optimisation. The experiment will study the impact of different initialisation strategies, including label (balanced or proportional) and feature (random noise, random subgraph, or k-Center graph~\cite{senerActiveLearningConvolutional2018}).

\textbf{RQ4:} How do different validation approaches impact the performance of graph condensation?

\textbf{Validation approaches} are essential for the graph condensation process. Validators, such as GNTK~\cite{duGraphNeuralTangent2019a} and GNNs are used to select the optimal condensed graph during training. The selected graphs are then compared to compare the effectiveness of the different validation approaches. 

\textbf{RQ5:} What is the effectiveness of the condensed graph for different GNN architectures?

\textbf{Cross-architecture transferability} is a key indicator of the generalisability of the condensed graph. To evaluate the cross-architecture transferability of the condensed graph, various model architectures are trained by the condensed graphs generated by different GC methods. 

\textbf{RQ6:} How effective are graph condensation methods when applied to continual graph learning?

\textbf{Continual graph learning (CGL)} is a widely studied downstream task for graph condensation. In CGL, a memory bank typically contains representative subgraphs of historical incoming graphs. However, considering the limitations of storage space and the effectiveness of subgraphs, condensed graphs may be a better choice. With the flexible design of \texttt{GCondenser}, GC methods can be easily applied to the continual graph learning task and used to evaluate performance in CGL.

%% file: sections/experiments.tex
\section{Experiments}

\begin{wraptable}{r}{0.5\textwidth}
\vspace{-0.2cm}
    \caption{Dataset statistics}\label{tab:datasets}
    \resizebox{1\linewidth}{!}{
    \begin{tabular}{ccccccc}\toprule
    Setting &Dataset &\#Nodes &\#Edges &\#Features &\#Classes  \\\midrule
     \multirow{5}{*}{Trans.} &CiteSeer &3,327 &4,732 &3,703 &6  \\
    &Cora &2,708 &5,429 &1,433 &7  \\
    &PubMed &19,717 &88,648 &500 &3  \\
    &ogbn-arxiv &169,343 &1,166,243 &128 &40 \\
    &ogbn-products &2,449,029 &61,859,140 &100 &47 \\
    \midrule
    \multirow{2}{*}{Ind.} &Flickr &89,250 &899,756 &500 &7 \\
    &Reddit &232,965 &57,307,946 &602 &41 \\
    \bottomrule
    \end{tabular}
}
\vspace{-0.3cm}
\end{wraptable}
\texttt{GCondenser} currently supports seven benchmark datasets with three small-scale networks (CiteSeer, Cora, and PubMed) and four larger graphs (ogbn-arxiv, ogbn-products, Flickr, and Reddit). The statistics are shown in Table~\ref{tab:datasets}. Flickr and Reddit are evaluated in the inductive (Ind.) setting, while the others are evaluated in the transductive (Trans.) setting. The details of transductive and inductive settings in graph condensation are shown in Figure~\ref{fig:setting} in Appendix~\ref{app:setting}.

\subsection{Baselines}
\textbf{Sampling approaches:} \textbf{Random} samples a random subgraph from the original graph dataset. \textbf{k-Center}~\cite{senerActiveLearningConvolutional2018} trains a GCN for several epochs to obtain the embeddings of the original graph, then samples the k-nearest nodes and retrieves the subgraph from the original graph.

\textbf{Gradient matching:} \textbf{GCond}~\cite{jinGraphCondensationGraph2022a} matches the gradients of training a GNN between the condensed graph and the original graph while training this GNN using the updated condensed graph. \textbf{GCondX} is the edge-free variant of GCond. \textbf{DosCond}~\cite{jinCondensingGraphsOneStep2022a} only matches the gradients of training a GNN between the condensed graph and the original graph on the first step of the GNN training. \textbf{DosCondX} is the edge-free variant of DosCond. \textbf{SGDD}~\cite{yangDoesGraphDistillation2023} replaces the feature-related MLP structure generator~\cite{jinGraphCondensationGraph2022a} with Graphon to mitigate the structure distribution shift on the spectral domain.

\textbf{Distribution matching:} \textbf{GCDM}~\cite{liuGraphCondensationReceptive2022} uses maximum mean discrepancy (MMD)~\cite{grettonKernelTwosampleTest2012} to measure the distances between original and condensed graph encoded by the same graph encoder during the GNN training using the condensed graph. \textbf{GCDMX} is the edge-free variant of GCDM. \textbf{DM}~\cite{liuCaTBalancedContinual2023a, liuPUMAEfficientContinual2023} uses one-step GCDM without learning any structures for efficiency.

\textbf{Trajectory matching:} \textbf{SFGC}~\cite{zhengStructurefreeGraphCondensation2023a} firstly generates training trajectories of training a GNN backbone using the original graph. For the condensation phase, a GNN with the same architecture is trained with the condensed graph, and the model parameters are matched to the generated model parameters. GEOM~\cite{geom} applies graph curriculum learning in the trajectory generation phase and uses expanding window matching and knowledge distillation to improve the quality of the condensed graph in the condensation phase.

\textbf{Eigenbasis matching:} \textbf{GDEM}~\cite{liuGraphCondensationEigenbasis2023} first performs eigendecomposition on the original dataset, then matches the synthetic bases with the real ones to reflect the structure and feature relations.

\subsection{Implementations}
\textbf{Dataset preprocessing:} Row feature normalisation is applied to the CiteSeer, Cora, and PubMed datasets, whilst standardisation is utilised on the Arxiv, Flickr, and Reddit datasets. The Products dataset continues to use the features processed by OGB~\cite{huOpenGraphBenchmark2020}.

\textbf{Initialisation:} The original label distribution and k-Center graph are utilised for initialisation by default. Different strategies for initialising the condensed graph are discussed in Section~\ref{sec:exp-init}.

\textbf{GNN usages:} SGC and GCN are applied as backbone models. The effectiveness of cross-architecture ability for validation and testing is discussed in Section~\ref{sec:exp-val} and~\ref{sec:exp-transferability}. 


\textbf{Hardware:} One NVIDIA L40 (42GB) or one NVIDIA V100 (32GB) GPU are used for experiment.

\textbf{Graph learning packages:} \texttt{GCondenser} refines baseline methods to support both popular graph learning packages, DGL and PyG, for convenience.

\begin{table}[!t]\centering
\caption{Overall node classification performance. For each dataset with different condensed graph sizes and backbone GNN models, the red and bold \textbf{\textcolor{red}{results}} indicate the best performance, while the blue and italic \underline{\textcolor{blue}{results}} indicate the runner-up. OOT stands for ‘Out of Time’ for 48 hours (e.g., GDEM runs out of time during the eigendecomposition phase).}
\label{tab:results}
\resizebox{\linewidth}{!}{
\input{./tables/overall_table}
}
\end{table}

\subsection{Overall Results (\textbf{RQ1})}
\label{sec:exp-overall}
Overall experiments are conducted with both SGC and GCN backbones for all datasets and baselines. For each dataset, three budgets are chosen as per~\cite{jinGraphCondensationGraph2022a}. The training, validation, and test GNN models are kept consistent for each experiment (e.g., SGC and GCN). Table~\ref{tab:results} shows the node classification accuracy of all baseline methods on seven datasets with three different budgets and two backbone models. For each condensed graph, different GC methods first define the hyperparameter candidates. A Bayesian hyperparameter sampler is then applied for faster hyperparameter tuning by monitoring the validation performance. The top three condensed graphs, based on validation performance for each target graph, are selected to report the average accuracy and standard deviation. To reproduce the experimental results, the choices of hyperparameters are provided in Appendix~\ref{app:rep}.

Experimental results show that all GC methods can achieve the highest or second-highest test accuracy in at least one setting. The trajectory-matching methods, SFGC and GEOM, show better performance on larger datasets with relatively small budgets. Specifically, SFGC has the best performance on the Arxiv dataset, while GDEM demonstrates superiority on the Products dataset. Surprisingly, even the simplest DM methods can effectively condense the graph. In addition, although GDEM performs well on small datasets, it encounters scalability issues with large graph datasets. Our findings are summarised as follows:

\textbf{Exploiting the potential of GC methods:} With effective validation approaches and comprehensive hyperparameter tuning, most methods achieve enhanced and comparable performance.

\textbf{Different backbones:} For \textbf{gradient matching} methods, using SGC as the backbone can produce more robust synthetic graphs, whereas GCN poses challenges due to its extra activation function, additional layers, and increased number of model parameters. For \textbf{distribution matching} methods, a suitable embedding space is crucial. The GCN backbone can encode nodes into a more expressive higher-dimensional space compared to SGC, which only has a linear fully connected layer. For \textbf{trajectory matching} methods, the more expressive GCN results in better synthetic graphs than SGC.

\textbf{Structure-free:} Generally, structure-free variants (e.g., GCondX, DosCondX and GCDMX) can obtain comparable results to condensed graphs with structure (e.g., GCond, DosCond, and GCDM). 

\begin{figure}[!h]
    \centering
    \includegraphics[width=1\linewidth]{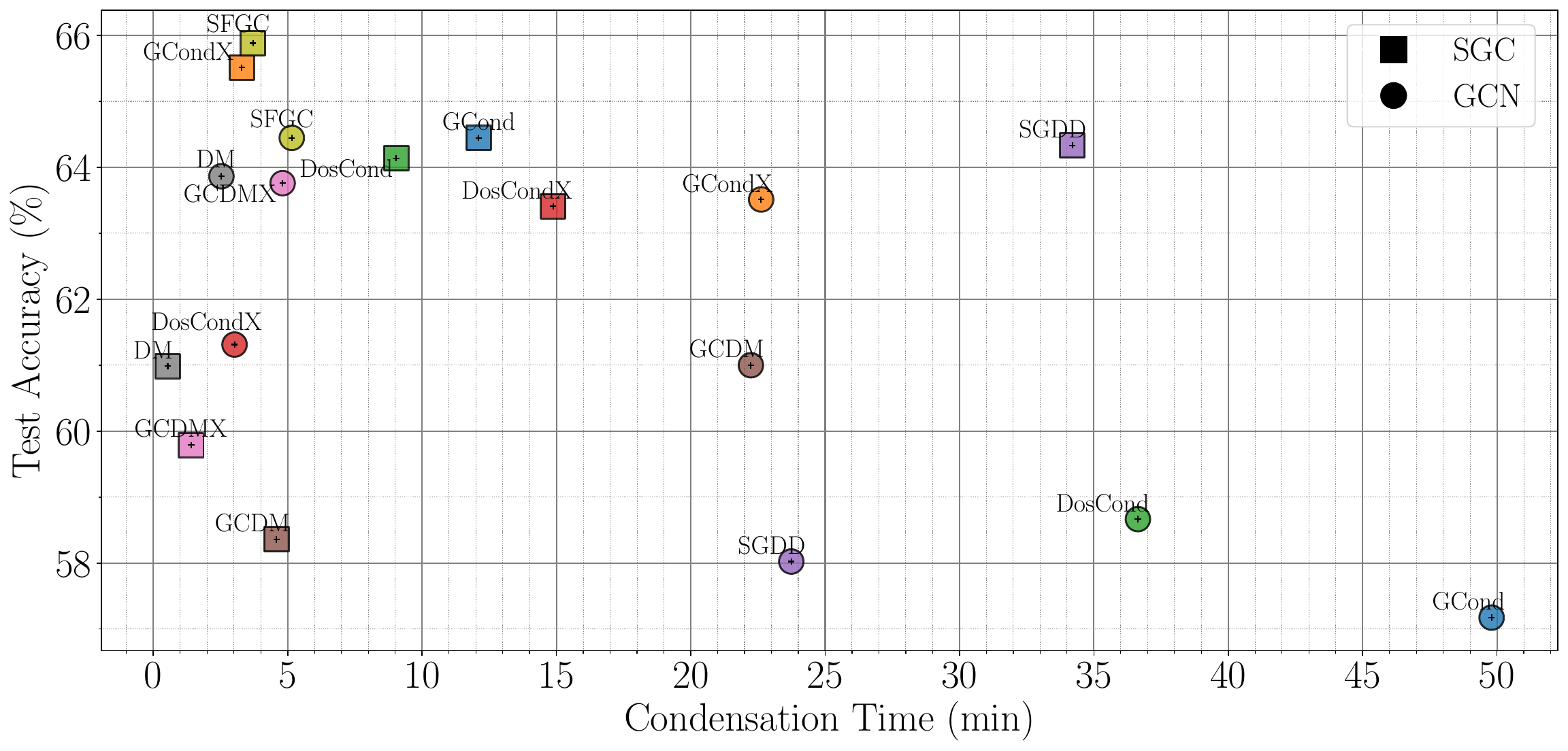}
    \caption{Test accuracy against condensation time for different GC methods on a 90-node condensed graph from the Arxiv dataset, with backbone models GCN and SGC. More results in Appendix~\ref{sec:efficiency-app}.}
    \label{fig:efficiency}
\end{figure}

\subsection{Condensation Efficiency (\textbf{RQ2})}
\label{sec:exp-efficiency}
Efficiency is a crucial aspect of evaluating different GC methods. In Figure~\ref{fig:efficiency}, the condensation time is compared against the model performance under the setting of 90-node for the Arxiv dataset. More results for different datasets are shown in Figure~\ref{fig:efficiency-citeseer}~$\sim$~\ref{fig:efficiency-reddit} of Appendix~\ref{sec:efficiency-app}.

Overall, methods only considering the feature matrix (e.g., GCondX, DosCondX, DM, SFGC) have a higher efficiency, while the methods considering both the feature matrix and the adjacency matrix (e.g., GCond, DosCond, SGDD, GCDM) need more condensation time. Moreover, gradient matching methods (e.g., GCond, GCondX, DosCond. DosCondX and SGDD) require more condensation time than distribution matching and trajectory matching methods. SGC as the backbone can improve the condensation efficiency but the model performance could drop slightly as in GCDM and DM.

\begin{wraptable}{r}{0.5\linewidth}\centering
    \vspace{-0cm}
    \caption{Storage space and the training time of offline trajectories in SFGC by training 200 GCNs.}
    \label{tab:traj_efficiency}
    \resizebox{1\linewidth}{!}{
    \begin{tabular}{lccccccc}\toprule
    &CiteSeer &Cora &PubMed &Arxiv &Products &Flickr &Reddit \\\midrule\midrule
    Storage (G) &72 &28 &9.8 &3.3 &3.1 &9.9 &13.3 \\
    Time (min) &58 &33 &50 &177 &2,820 &184 &1,961 \\
    \bottomrule
    \end{tabular}
\vspace{-0cm}
}
\end{wraptable}
 For SFGC, extra storage space and time are needed for trajectory generations, as in Table~\ref{tab:traj_efficiency}. By training 200 offline expert GCNs, the overall condensation cost of SFGC will increase significantly, which would lead to either storage issues or a slow condensation process, which is not applicable for online downstream applications, such as continual graph learning in Section~\ref{sec:exp-cgl}.

\subsection{Condensed Graph Initialisation (RQ3)}
\label{sec:exp-init}
\begin{wraptable}{r}{0.5\linewidth}
\vspace{-0.5cm}
    \caption{Condensation of Arxiv into 90 nodes with different condensed graph initialisation.}\label{tab:init}
    \resizebox{1\linewidth}{!}{\input{./tables/init}}
\vspace{-0.2cm}
\end{wraptable}
Different initialisation strategies for the condensed graphs also affect the optimisation and output quality of GC methods. Stable optimisation and consistent condensed graph quality across different initialisations are desired for practical applications. Table~\ref{tab:init} presents the performance of three methods (GCond, DosCond, and DM) on the Arxiv dataset for two label distributions (balanced and proportional to original) and three feature initialisation methods (random noise, random subgraph, and k-Center graph). The results show that maintaining the original label distribution leads to better performance than using a balanced label distribution. This is because, for graph data, there is generally a class imbalance issue with a large number of classes (40 classes for Arxiv). Different label initialisations directly constrain the node budget for each class. For instance, Figure~\ref{fig:label_distribution} in Appendix~\ref{app:label_distribution} illustrates how class sizes vary in the condensed graph under different label initialisations. Keeping the proportion of labels in the condensed graph can largely maintain the performance of the classes with more nodes. While for the feature initialisation, different strategies achieve comparable results across different GC methods.

\subsection{Impact of Different Validators (\textbf{RQ4})}
\label{sec:exp-val}
The validator selection for the graph condensation process is essential. The validator can select a convincing graph during each condensation process and guide the hyperparameter sampler to quickly and accurately find the better hyperparameter combination. For the reliable quality of the selected graph and the effectiveness of hyperparameter search, a performance consistency of the validator on the validation set and test set is expected. GCN, SGC and GNTK as validators can serve the graph condensation process to find the optimal graph that optimises the validator to perform best on the validation set. Table~\ref{tab:evaluator} shows the performance of the GCN trained with different validators and GC methods and the relative convergence time compared with GCN. The results indicate that GNN validators (e.g., GCN and SGC) are effective in assessing the quality of the condensed graph. GCN produces better results but requires more time, while SGC, which caches the aggregated features across the graph, requires less time but sacrifices some performance. GNTK is the most efficient validator, but it is not as reliable compared to GNN validators.
\label{sec:exp-evaluator}

\begin{table}[!t]
    \centering
    \caption{Performance of different validators on Cora, with a budget of 35-node and GCN as the backbone. The test accuracy (\%) and the relative condensation time to the GCN validator are shown.}
    \resizebox{\linewidth}{!}{
    \input{./tables/cora-35-evaluator-test-acc}

    }
    \label{tab:evaluator}
\end{table}

\subsection{Cross-architecture Transferability (\textbf{RQ5})}
\label{sec:exp-transferability}
\begin{figure}[!h]
    \centering
    \includegraphics[width=1\textwidth]{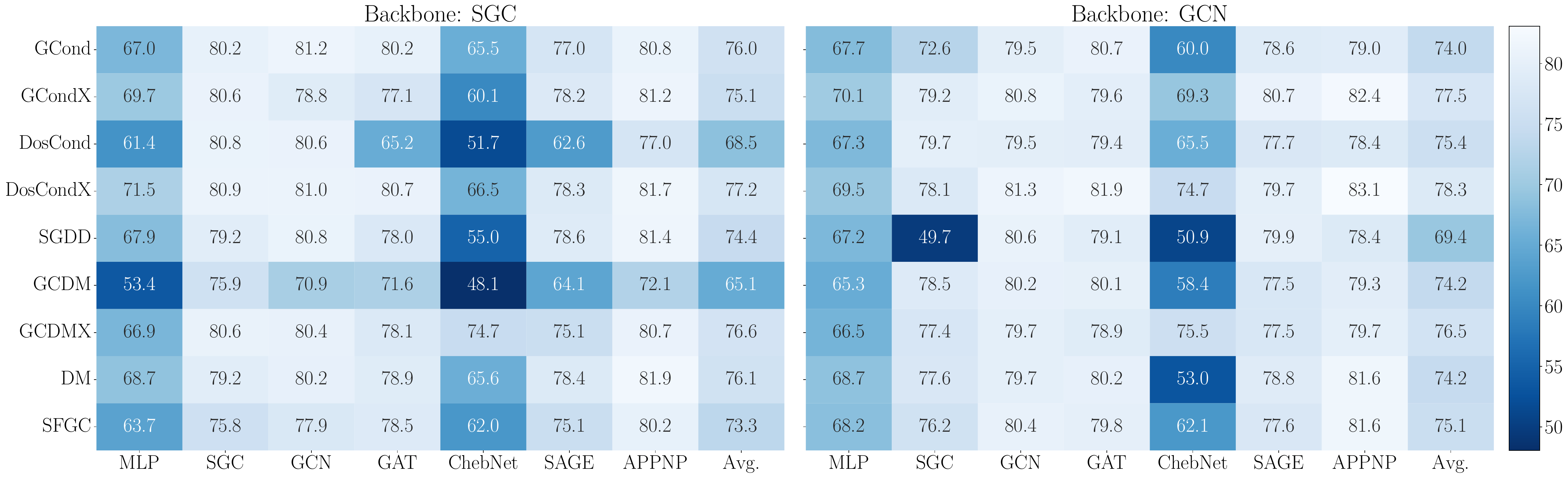}
    \caption{Transferability of condensed graphs for Cora with budget 35. More results in Appendix~\ref{sec:exp-transferability-app}.}
    \label{fig:transferability}
\end{figure}

To test the cross-architecture transferability of optimal condensed graphs from Section~\ref{sec:exp-overall} After obtaining the selected optimal condensed graph. \texttt{GCondenser} can use the condensed graph to train different GNNs and evaluate the performance of the GNNs on the test set. The benchmark provides a set of architectures, including MLP,  SGC~\cite{wuSimplifyingGraphConvolutional2019a}, GCN~\cite{kipfSemisupervisedClassificationGraph2017a}, GAT~\cite{velickovicGraphAttentionNetworks2018a}, ChebNet~\cite{defferrardConvolutionalNeuralNetworks2016a}, SAGE~\cite{hamiltonInductiveRepresentationLearning2017}, APPNP~\cite{klicperaPredictThenPropagate2019a}. It is worth noting that for MLP evaluation, graph structure information is avoided in both the training and inference phases. Figure~\ref{fig:transferability} shows the cross-architecture transferability of the 30-node graph condensed from the CiteSeer dataset by different GC methods. The results show that the DosCondX and GCondX methods with the GCN backbone have better average performance. More cross-architecture evaluations with different datasets are shown in Figure~\ref{fig:transferability-citeseer}~$\sim$~\ref{fig:transferability-reddit} of Appendix~\ref{sec:exp-transferability-app}.

\subsection{Continual Graph Learning (RQ6)}
\label{sec:exp-cgl}
To explore the effectiveness of different graph condensation methods in downstream tasks, continual graph learning (CGL) is employed for evaluating the condensed graphs. Three representative methods are selected for comparison: GCond for multi-step matching with structure learning, DosCond for one-step matching with structures learning, and DM for one-step matching without structure learning. These three methods are compared against the random subgraph method, the k-Center graph method, and the whole graph upper-bound. It is worth noting that SFGC, as an offline algorithm, is not suitable to apply in continual graph learning, where online updates are required. The Condense and Train (CaT) framework~\cite{liuCaTBalancedContinual2023a} under the class incremental learning (class-IL) setting is applied. CiteSeer, Cora, Arxiv, Products, Flickr, and Reddit datasets are divided into a series of incoming subgraphs, each with two new classes. If a subgraph contains only one class, this subgraph should be removed from the streaming data. PubMed is not evaluated as it only contains three classes. Offline methods, such as SFGC, GDEM, and GEOM, are also not evaluated. Due to the running time of SGDD, a large number of tasks would be time-consuming. Therefore, SGDD is not evaluated in the CGL.

\begin{figure}[!h]
    \centering
    \includegraphics[width=\linewidth]{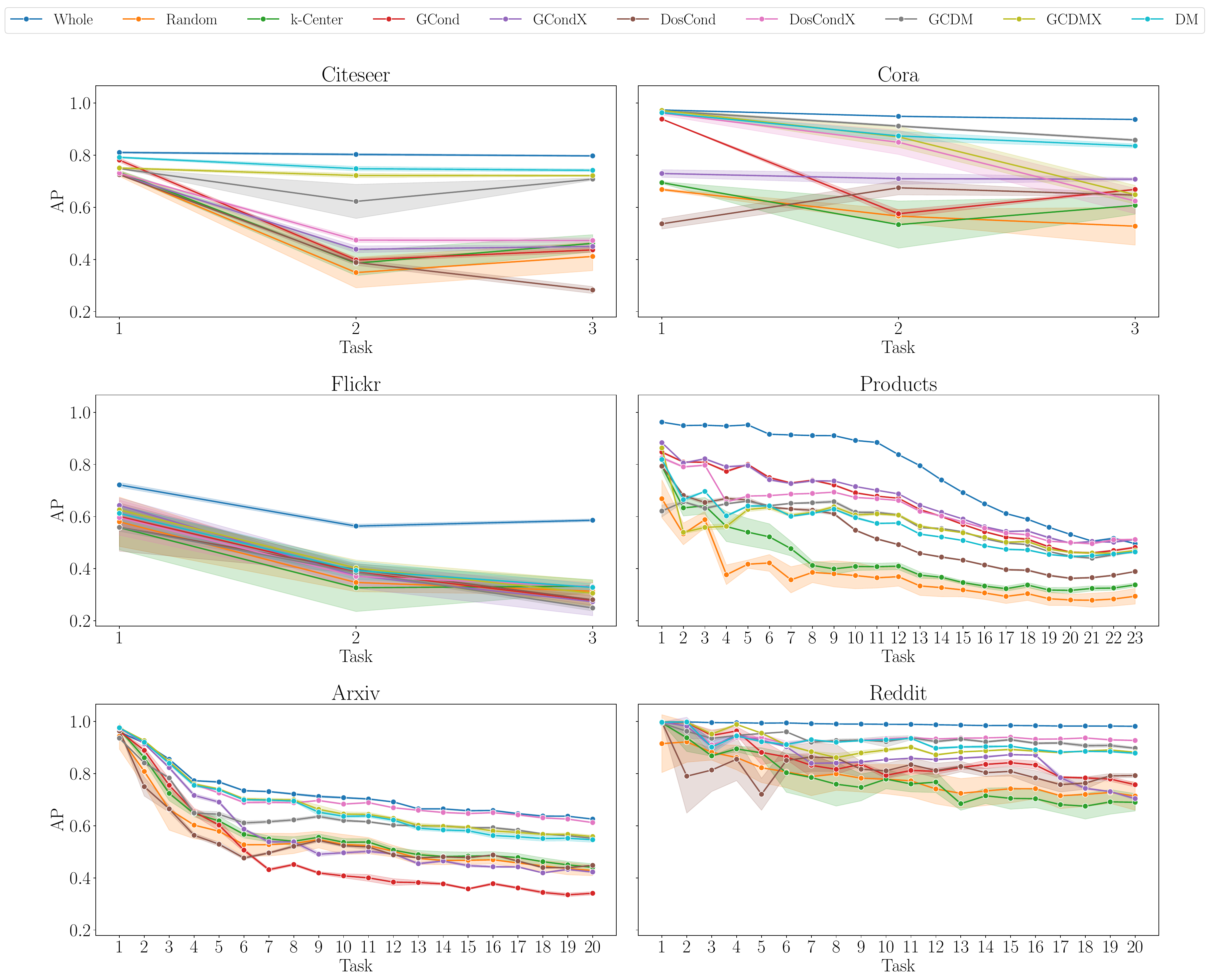}
    \caption{GC methods on baseline datasets under the continual graph learning setting.}
    \label{fig:cgl-supp}
\end{figure}

Figure~\ref{fig:cgl-supp} shows that in smaller datasets like CiteSeer and Cora, distribution-matching methods perform better due to the good adaptability of the condensed graphs to the continually learned model. DosCondX can even match the performance of training on the whole graph in long series tasks. However, most methods failed on the Flickr dataset, indicating that this dataset poses challenges for maintaining historical knowledge on the condensed graphs in the CGL setting.

%% file: tables/overall_table.tex
\begin{tabular}{c|c|ccc|cc|ccccc|ccc|cc|c|c}\toprule
\multicolumn{2}{c|}{\multirow{2}{*}[-1ex]{Dataset}} &\multirow{2}{*}[-1ex]{GNN} &\multirow{2}{*}[-1ex]{Size} &\multirow{2}{*}[-1ex]{Ratio} &\multicolumn{2}{c|}{Traditional} &\multicolumn{5}{c|}{Gradient} &\multicolumn{3}{c|}{Distribution} &\multicolumn{2}{c|}{Trajectory} &Eigenbasis &\multirow{2}{*}[-1ex]{Whole}\\
\cmidrule{6-18}
 \multicolumn{2}{c|}{\multirow{2}{*}{}}& & & &Random &k-Center &GCond &GCondX &DosCond &DosCondX &SGDD &GCDM &GCDMX &DM &SFGC &GEOM &GDEM &\\\midrule\midrule
\multirow{24}{*}[-2.5cm]{\rotatebox{90}{Transductive}} &\multirow{6}{*}[-0.3cm]{\rotatebox{90}{CiteSeer}} & \multirow{3}{*}[-0.4ex]{SGC} & 30 &0.9\% &53.8\textsubscript{\textnormal{$\pm$}\textnormal{0.1}} &52.7\textsubscript{\textnormal{$\pm$}\textnormal{0.0}} &\textbf{\textcolor{red}{71.9\textsubscript{\textnormal{$\pm$}\textnormal{0.6}}}} &65.3\textsubscript{\textnormal{$\pm$}\textnormal{0.1}} &\underline{\textcolor{blue}{71.1\textsubscript{\textnormal{$\pm$}\textnormal{0.9}}}} &64.3\textsubscript{\textnormal{$\pm$}\textnormal{0.4}} &71.1\textsubscript{\textnormal{$\pm$}\textnormal{0.1}} &66.0\textsubscript{\textnormal{$\pm$}\textnormal{2.2}} &70.1\textsubscript{\textnormal{$\pm$}\textnormal{0.8}} &64.2\textsubscript{\textnormal{$\pm$}\textnormal{8.5}} &65.2\textsubscript{\textnormal{$\pm$}\textnormal{0.3}} &60.1\textsubscript{\textnormal{$\pm$}\textnormal{0.2}} &70.8\textsubscript{\textnormal{$\pm$}\textnormal{0.5}} &\multirow{3}{*}[-0.4ex]{70.3±1.0} \\
 & &  & 60 &1.8\% &61.9\textsubscript{\textnormal{$\pm$}\textnormal{0.0}} &66.8\textsubscript{\textnormal{$\pm$}\textnormal{0.0}} &71.0\textsubscript{\textnormal{$\pm$}\textnormal{0.6}} &70.9\textsubscript{\textnormal{$\pm$}\textnormal{0.3}} &\underline{\textcolor{blue}{72.0\textsubscript{\textnormal{$\pm$}\textnormal{1.1}}}} &70.6\textsubscript{\textnormal{$\pm$}\textnormal{0.1}} &69.9\textsubscript{\textnormal{$\pm$}\textnormal{0.1}} &66.7\textsubscript{\textnormal{$\pm$}\textnormal{0.0}} &65.8\textsubscript{\textnormal{$\pm$}\textnormal{1.3}} &65.7\textsubscript{\textnormal{$\pm$}\textnormal{2.5}} &67.0\textsubscript{\textnormal{$\pm$}\textnormal{0.8}} &65.2\textsubscript{\textnormal{$\pm$}\textnormal{0.2}}  &\textbf{\textcolor{red}{72.3\textsubscript{\textnormal{$\pm$}\textnormal{0.6}}}} &  \\
 & &  & 120 &3.6\% &68.1\textsubscript{\textnormal{$\pm$}\textnormal{0.0}} &68.1\textsubscript{\textnormal{$\pm$}\textnormal{0.0}} &\underline{\textcolor{blue}{72.5\textsubscript{\textnormal{$\pm$}\textnormal{1.2}}}} &71.5\textsubscript{\textnormal{$\pm$}\textnormal{0.0}} &70.6\textsubscript{\textnormal{$\pm$}\textnormal{0.2}} &71.8\textsubscript{\textnormal{$\pm$}\textnormal{0.3}} &70.8\textsubscript{\textnormal{$\pm$}\textnormal{0.8}} &69.1\textsubscript{\textnormal{$\pm$}\textnormal{1.2}} &68.1\textsubscript{\textnormal{$\pm$}\textnormal{0.2}} &69.7\textsubscript{\textnormal{$\pm$}\textnormal{0.1}} &68.8\textsubscript{\textnormal{$\pm$}\textnormal{0.2}} &67.7\textsubscript{\textnormal{$\pm$}\textnormal{0.3}} &\textbf{\textcolor{red}{72.8\textsubscript{\textnormal{$\pm$}\textnormal{0.4}}}} &  \\
\cmidrule{3-19}
 & & \multirow{3}{*}[-0.4ex]{GCN} & 30 &0.9\% &58.9\textsubscript{\textnormal{$\pm$}\textnormal{0.0}} &65.0\textsubscript{\textnormal{$\pm$}\textnormal{0.0}} &46.3\textsubscript{\textnormal{$\pm$}\textnormal{7.0}} &66.7\textsubscript{\textnormal{$\pm$}\textnormal{4.2}} &69.2\textsubscript{\textnormal{$\pm$}\textnormal{0.5}} &67.7\textsubscript{\textnormal{$\pm$}\textnormal{1.2}} &70.6\textsubscript{\textnormal{$\pm$}\textnormal{1.5}} &71.2\textsubscript{\textnormal{$\pm$}\textnormal{0.8}} &72.8\textsubscript{\textnormal{$\pm$}\textnormal{0.3}} &\textbf{\textcolor{red}{72.6\textsubscript{\textnormal{$\pm$}\textnormal{0.6}}}} &69.7\textsubscript{\textnormal{$\pm$}\textnormal{0.3}} &69.6\textsubscript{\textnormal{$\pm$}\textnormal{0.6}} &\underline{\textcolor{blue}{71.7\textsubscript{\textnormal{$\pm$}\textnormal{0.3}}}} &\multirow{3}{*}[-0.4ex]{71.4±0.5} \\
 & &  & 60 &1.8\% &62.6\textsubscript{\textnormal{$\pm$}\textnormal{0.0}} &67.8\textsubscript{\textnormal{$\pm$}\textnormal{0.0}} &54.2\textsubscript{\textnormal{$\pm$}\textnormal{3.9}} &69.6\textsubscript{\textnormal{$\pm$}\textnormal{0.4}} &70.4\textsubscript{\textnormal{$\pm$}\textnormal{1.7}} &70.1\textsubscript{\textnormal{$\pm$}\textnormal{0.2}} &71.5\textsubscript{\textnormal{$\pm$}\textnormal{0.7}} &71.9\textsubscript{\textnormal{$\pm$}\textnormal{0.7}} &71.7\textsubscript{\textnormal{$\pm$}\textnormal{0.2}} &\underline{\textcolor{blue}{72.2\textsubscript{\textnormal{$\pm$}\textnormal{0.7}}}} &69.4\textsubscript{\textnormal{$\pm$}\textnormal{0.0}} &67.5\textsubscript{\textnormal{$\pm$}\textnormal{0.9}} &\textbf{\textcolor{red}{72.7\textsubscript{\textnormal{$\pm$}\textnormal{0.6}}}} &  \\
 & &  & 120 &3.6\% &69.6\textsubscript{\textnormal{$\pm$}\textnormal{0.0}} &69.4\textsubscript{\textnormal{$\pm$}\textnormal{0.0}} &70.7\textsubscript{\textnormal{$\pm$}\textnormal{0.7}} &70.9\textsubscript{\textnormal{$\pm$}\textnormal{0.2}} &47.3\textsubscript{\textnormal{$\pm$}\textnormal{7.3}} &70.4\textsubscript{\textnormal{$\pm$}\textnormal{0.2}} &71.0\textsubscript{\textnormal{$\pm$}\textnormal{0.7}} &72.3\textsubscript{\textnormal{$\pm$}\textnormal{1.3}} &\underline{\textcolor{blue}{72.5\textsubscript{\textnormal{$\pm$}\textnormal{0.5}}}} &72.4\textsubscript{\textnormal{$\pm$}\textnormal{0.1}} &69.8\textsubscript{\textnormal{$\pm$}\textnormal{0.5}} &72.1\textsubscript{\textnormal{$\pm$}\textnormal{1.0}} &\textbf{\textcolor{red}{73.4\textsubscript{\textnormal{$\pm$}\textnormal{0.4}}}} &  \\
\cmidrule{2-19}
& \multirow{6}{*}[-2ex]{\rotatebox{90}{Cora}} & \multirow{3}{*}[-0.4ex]{SGC} & 35 &1.3\% &59.3\textsubscript{\textnormal{$\pm$}\textnormal{0.0}} &63.8\textsubscript{\textnormal{$\pm$}\textnormal{0.0}} &80.6\textsubscript{\textnormal{$\pm$}\textnormal{0.1}} &\underline{\textcolor{blue}{80.6\textsubscript{\textnormal{$\pm$}\textnormal{0.2}}}} &80.6\textsubscript{\textnormal{$\pm$}\textnormal{0.1}} &\textbf{\textcolor{red}{80.8\textsubscript{\textnormal{$\pm$}\textnormal{0.1}}}} &62.4\textsubscript{\textnormal{$\pm$}\textnormal{5.5}} &77.0\textsubscript{\textnormal{$\pm$}\textnormal{0.4}} &79.9\textsubscript{\textnormal{$\pm$}\textnormal{0.1}} &77.0\textsubscript{\textnormal{$\pm$}\textnormal{0.5}} &73.8\textsubscript{\textnormal{$\pm$}\textnormal{1.5}} &69.2\textsubscript{\textnormal{$\pm$}\textnormal{1.2}} &71.0\textsubscript{\textnormal{$\pm$}\textnormal{0.6}} &\multirow{3}{*}[-0.4ex]{79.2±0.6} \\
 & &  & 70 &2.6\% &70.7\textsubscript{\textnormal{$\pm$}\textnormal{0.0}} &70.3\textsubscript{\textnormal{$\pm$}\textnormal{0.0}} &\textbf{\textcolor{red}{81.0\textsubscript{\textnormal{$\pm$}\textnormal{0.2}}}} &79.0\textsubscript{\textnormal{$\pm$}\textnormal{0.3}} &80.3\textsubscript{\textnormal{$\pm$}\textnormal{0.5}} &79.8\textsubscript{\textnormal{$\pm$}\textnormal{0.2}} &\underline{\textcolor{blue}{80.8\textsubscript{\textnormal{$\pm$}\textnormal{0.4}}}} &78.9\textsubscript{\textnormal{$\pm$}\textnormal{1.0}} &79.7\textsubscript{\textnormal{$\pm$}\textnormal{0.3}} &78.0\textsubscript{\textnormal{$\pm$}\textnormal{1.6}} &77.5\textsubscript{\textnormal{$\pm$}\textnormal{0.1}} &69.6\textsubscript{\textnormal{$\pm$}\textnormal{1.5}} &75.4\textsubscript{\textnormal{$\pm$}\textnormal{0.5}} &  \\
 & &  & 140 &5.2\% &77.0\textsubscript{\textnormal{$\pm$}\textnormal{0.0}} &77.1\textsubscript{\textnormal{$\pm$}\textnormal{0.0}} &80.9\textsubscript{\textnormal{$\pm$}\textnormal{0.4}} &\underline{\textcolor{blue}{81.2\textsubscript{\textnormal{$\pm$}\textnormal{0.3}}}} &80.8\textsubscript{\textnormal{$\pm$}\textnormal{0.4}} &80.8\textsubscript{\textnormal{$\pm$}\textnormal{0.5}} &\textbf{\textcolor{red}{81.4\textsubscript{\textnormal{$\pm$}\textnormal{0.4}}}} &77.9\textsubscript{\textnormal{$\pm$}\textnormal{0.7}} &80.1\textsubscript{\textnormal{$\pm$}\textnormal{0.2}} &79.6\textsubscript{\textnormal{$\pm$}\textnormal{0.6}} &79.2\textsubscript{\textnormal{$\pm$}\textnormal{0.1}} &77.3\textsubscript{\textnormal{$\pm$}\textnormal{0.1}}  &79.5\textsubscript{\textnormal{$\pm$}\textnormal{0.8}} &  \\
\cmidrule{3-19}
 & & \multirow{3}{*}[-0.4ex]{GCN} & 35 &1.3\% &63.9\textsubscript{\textnormal{$\pm$}\textnormal{0.1}} &66.5\textsubscript{\textnormal{$\pm$}\textnormal{0.0}} &\underline{\textcolor{blue}{80.5\textsubscript{\textnormal{$\pm$}\textnormal{0.4}}}} &79.8\textsubscript{\textnormal{$\pm$}\textnormal{1.4}} &72.0\textsubscript{\textnormal{$\pm$}\textnormal{8.7}} &80.1\textsubscript{\textnormal{$\pm$}\textnormal{0.9}} &\textbf{\textcolor{red}{80.5\textsubscript{\textnormal{$\pm$}\textnormal{0.4}}}} &78.9\textsubscript{\textnormal{$\pm$}\textnormal{0.8}} &79.1\textsubscript{\textnormal{$\pm$}\textnormal{0.9}} &79.2\textsubscript{\textnormal{$\pm$}\textnormal{0.2}} &79.6\textsubscript{\textnormal{$\pm$}\textnormal{0.2}} &80.3\textsubscript{\textnormal{$\pm$}\textnormal{1.1}} &68.0\textsubscript{\textnormal{$\pm$}\textnormal{0.1}} & \multirow{3}{*}[-0.4ex]{81.7±0.9} \\
 & &  & 70 &2.6\% &73.0\textsubscript{\textnormal{$\pm$}\textnormal{0.0}} &71.6\textsubscript{\textnormal{$\pm$}\textnormal{0.0}} &78.1\textsubscript{\textnormal{$\pm$}\textnormal{3.6}} &80.6\textsubscript{\textnormal{$\pm$}\textnormal{0.9}} &79.6\textsubscript{\textnormal{$\pm$}\textnormal{0.7}} &\underline{\textcolor{blue}{81.1\textsubscript{\textnormal{$\pm$}\textnormal{0.3}}}} &\textbf{\textcolor{red}{81.2\textsubscript{\textnormal{$\pm$}\textnormal{0.6}}}} &79.4\textsubscript{\textnormal{$\pm$}\textnormal{0.6}} &80.5\textsubscript{\textnormal{$\pm$}\textnormal{0.3}} &79.6\textsubscript{\textnormal{$\pm$}\textnormal{0.3}} &79.5\textsubscript{\textnormal{$\pm$}\textnormal{0.1}} &81.5\textsubscript{\textnormal{$\pm$}\textnormal{0.8}} &72.8\textsubscript{\textnormal{$\pm$}\textnormal{0.8}} &  \\
 & &  & 140 &5.2\% &77.1\textsubscript{\textnormal{$\pm$}\textnormal{0.0}} &76.6\textsubscript{\textnormal{$\pm$}\textnormal{0.0}} &80.2\textsubscript{\textnormal{$\pm$}\textnormal{1.7}} &\textbf{\textcolor{red}{81.5\textsubscript{\textnormal{$\pm$}\textnormal{0.1}}}} &80.5\textsubscript{\textnormal{$\pm$}\textnormal{0.7}} &\underline{\textcolor{blue}{80.6\textsubscript{\textnormal{$\pm$}\textnormal{0.2}}}} &79.9\textsubscript{\textnormal{$\pm$}\textnormal{1.6}} &79.9\textsubscript{\textnormal{$\pm$}\textnormal{0.2}} &80.2\textsubscript{\textnormal{$\pm$}\textnormal{0.5}} &79.9\textsubscript{\textnormal{$\pm$}\textnormal{0.3}} &80.1\textsubscript{\textnormal{$\pm$}\textnormal{0.6}} &82.2\textsubscript{\textnormal{$\pm$}\textnormal{0.4}} &77.4\textsubscript{\textnormal{$\pm$}\textnormal{0.6}} &  \\
\cmidrule{2-19}
& \multirow{6}{*}[-1.5ex]{\rotatebox{90}{PubMed}} &\multirow{3}{*}[-0.4ex]{SGC} & 15 &0.08\% &67.8\textsubscript{\textnormal{$\pm$}\textnormal{0.2}} &70.5\textsubscript{\textnormal{$\pm$}\textnormal{0.1}} &75.9\textsubscript{\textnormal{$\pm$}\textnormal{0.7}} &\textbf{\textcolor{red}{77.3\textsubscript{\textnormal{$\pm$}\textnormal{0.2}}}} &74.2\textsubscript{\textnormal{$\pm$}\textnormal{1.1}} &75.7\textsubscript{\textnormal{$\pm$}\textnormal{0.5}} &\underline{\textcolor{blue}{76.4\textsubscript{\textnormal{$\pm$}\textnormal{0.9}}}} &73.3\textsubscript{\textnormal{$\pm$}\textnormal{1.2}} &74.0\textsubscript{\textnormal{$\pm$}\textnormal{0.3}} &72.1\textsubscript{\textnormal{$\pm$}\textnormal{0.9}} &73.9\textsubscript{\textnormal{$\pm$}\textnormal{0.5}} &73.8\textsubscript{\textnormal{$\pm$}\textnormal{0.3}} &73.8\textsubscript{\textnormal{$\pm$}\textnormal{0.6}} & \multirow{3}{*}[-0.4ex]{76.9±0.1} \\
 & &  & 30 &0.15\% &72.5\textsubscript{\textnormal{$\pm$}\textnormal{0.2}} &75.8\textsubscript{\textnormal{$\pm$}\textnormal{0.0}} &75.2\textsubscript{\textnormal{$\pm$}\textnormal{0.0}} &77.1\textsubscript{\textnormal{$\pm$}\textnormal{0.2}} &77.2\textsubscript{\textnormal{$\pm$}\textnormal{0.1}} &77.0\textsubscript{\textnormal{$\pm$}\textnormal{0.1}} &\underline{\textcolor{blue}{78.0\textsubscript{\textnormal{$\pm$}\textnormal{0.3}}}} &74.7\textsubscript{\textnormal{$\pm$}\textnormal{0.6}} &75.2\textsubscript{\textnormal{$\pm$}\textnormal{0.7}} &75.1\textsubscript{\textnormal{$\pm$}\textnormal{0.6}} &75.8\textsubscript{\textnormal{$\pm$}\textnormal{0.2}} &77.4\textsubscript{\textnormal{$\pm$}\textnormal{0.4}} &\textbf{\textcolor{red}{78.7\textsubscript{\textnormal{$\pm$}\textnormal{0.4}}}} & \\
 & &  & 60 &0.3\% &75.6\textsubscript{\textnormal{$\pm$}\textnormal{0.0}} &75.7\textsubscript{\textnormal{$\pm$}\textnormal{0.0}} &75.7\textsubscript{\textnormal{$\pm$}\textnormal{0.0}} &\underline{\textcolor{blue}{76.8\textsubscript{\textnormal{$\pm$}\textnormal{0.1}}}} &75.5\textsubscript{\textnormal{$\pm$}\textnormal{0.2}} &75.5\textsubscript{\textnormal{$\pm$}\textnormal{0.0}} &76.1\textsubscript{\textnormal{$\pm$}\textnormal{0.1}} &76.5\textsubscript{\textnormal{$\pm$}\textnormal{1.1}} &76.3\textsubscript{\textnormal{$\pm$}\textnormal{0.2}} &75.8\textsubscript{\textnormal{$\pm$}\textnormal{0.0}} &75.8\textsubscript{\textnormal{$\pm$}\textnormal{0.0}} &75.8\textsubscript{\textnormal{$\pm$}\textnormal{0.4}} &\textbf{\textcolor{red}{79.1\textsubscript{\textnormal{$\pm$}\textnormal{0.1}}}} &  \\
\cmidrule{3-19}
 & & \multirow{3}{*}[-0.4ex]{GCN} & 15 &0.08\% &69.8\textsubscript{\textnormal{$\pm$}\textnormal{0.1}} &72.1\textsubscript{\textnormal{$\pm$}\textnormal{0.1}} &67.6\textsubscript{\textnormal{$\pm$}\textnormal{10.4}} &77.8\textsubscript{\textnormal{$\pm$}\textnormal{0.3}} &75.7\textsubscript{\textnormal{$\pm$}\textnormal{0.7}} &75.7\textsubscript{\textnormal{$\pm$}\textnormal{0.1}} &76.7\textsubscript{\textnormal{$\pm$}\textnormal{1.1}} &75.9\textsubscript{\textnormal{$\pm$}\textnormal{0.6}} &77.1\textsubscript{\textnormal{$\pm$}\textnormal{0.3}} &76.0\textsubscript{\textnormal{$\pm$}\textnormal{0.7}} &\underline{\textcolor{blue}{78.4\textsubscript{\textnormal{$\pm$}\textnormal{0.1}}}} &\textbf{\textcolor{red}{80.1\textsubscript{\textnormal{$\pm$}\textnormal{0.3}}}} &73.3\textsubscript{\textnormal{$\pm$}\textnormal{0.6}} & \multirow{3}{*}[-0.4ex]{79.3±0.3} \\
 & &  & 30 &0.15\% &73.7\textsubscript{\textnormal{$\pm$}\textnormal{0.1}} &76.4\textsubscript{\textnormal{$\pm$}\textnormal{0.0}} &74.6\textsubscript{\textnormal{$\pm$}\textnormal{0.8}} &78.0\textsubscript{\textnormal{$\pm$}\textnormal{0.5}} &76.8\textsubscript{\textnormal{$\pm$}\textnormal{0.2}} &\underline{\textcolor{blue}{78.6\textsubscript{\textnormal{$\pm$}\textnormal{0.2}}}} &78.5\textsubscript{\textnormal{$\pm$}\textnormal{0.4}} &77.4\textsubscript{\textnormal{$\pm$}\textnormal{0.4}} &76.8\textsubscript{\textnormal{$\pm$}\textnormal{0.6}} &77.5\textsubscript{\textnormal{$\pm$}\textnormal{0.1}} &78.1\textsubscript{\textnormal{$\pm$}\textnormal{0.4}} &\textbf{\textcolor{red}{79.7\textsubscript{\textnormal{$\pm$}\textnormal{0.3}}}} &78.3\textsubscript{\textnormal{$\pm$}\textnormal{0.8}} &  \\
 & &  & 60 &0.3\% &78.0\textsubscript{\textnormal{$\pm$}\textnormal{0.0}} &78.2\textsubscript{\textnormal{$\pm$}\textnormal{0.0}} &77.2\textsubscript{\textnormal{$\pm$}\textnormal{0.7}} &78.0\textsubscript{\textnormal{$\pm$}\textnormal{0.1}} &77.3\textsubscript{\textnormal{$\pm$}\textnormal{1.2}} &78.0\textsubscript{\textnormal{$\pm$}\textnormal{0.1}} &78.0\textsubscript{\textnormal{$\pm$}\textnormal{1.1}} &77.6\textsubscript{\textnormal{$\pm$}\textnormal{0.4}} &78.1\textsubscript{\textnormal{$\pm$}\textnormal{0.3}} &78.0\textsubscript{\textnormal{$\pm$}\textnormal{0.2}} &78.5\textsubscript{\textnormal{$\pm$}\textnormal{0.5}} &\textbf{\textcolor{red}{79.5\textsubscript{\textnormal{$\pm$}\textnormal{0.4}}}} &\underline{\textcolor{blue}{78.8\textsubscript{\textnormal{$\pm$}\textnormal{0.3}}}} &  \\
\cmidrule{2-19}
& \multirow{6}{*}[-1.5ex]{\rotatebox{90}{Arxiv}} & \multirow{3}{*}[-0.4ex]{SGC} & 90 &0.05\% &45.6\textsubscript{\textnormal{$\pm$}\textnormal{0.2}} &51.8\textsubscript{\textnormal{$\pm$}\textnormal{0.2}} &65.5\textsubscript{\textnormal{$\pm$}\textnormal{0.0}} &\underline{\textcolor{blue}{66.0\textsubscript{\textnormal{$\pm$}\textnormal{0.2}}}} &62.7\textsubscript{\textnormal{$\pm$}\textnormal{0.6}} &61.6\textsubscript{\textnormal{$\pm$}\textnormal{0.3}} &64.5\textsubscript{\textnormal{$\pm$}\textnormal{0.9}} &60.8\textsubscript{\textnormal{$\pm$}\textnormal{0.1}} &59.8\textsubscript{\textnormal{$\pm$}\textnormal{0.3}} &61.0\textsubscript{\textnormal{$\pm$}\textnormal{0.2}} &\textbf{\textcolor{red}{66.1\textsubscript{\textnormal{$\pm$}\textnormal{0.2}}}} &62.0\textsubscript{\textnormal{$\pm$}\textnormal{0.5}} &OOT & \multirow{3}{*}[-0.4ex]{68.8±0.0} \\
 & &  & 454 &0.25\% &55.2\textsubscript{\textnormal{$\pm$}\textnormal{0.0}} &58.2\textsubscript{\textnormal{$\pm$}\textnormal{0.0}} &\underline{\textcolor{blue}{66.5\textsubscript{\textnormal{$\pm$}\textnormal{0.5}}}} &66.4\textsubscript{\textnormal{$\pm$}\textnormal{0.1}} &63.7\textsubscript{\textnormal{$\pm$}\textnormal{0.2}} &63.8\textsubscript{\textnormal{$\pm$}\textnormal{0.1}} &66.4\textsubscript{\textnormal{$\pm$}\textnormal{0.3}} &62.7\textsubscript{\textnormal{$\pm$}\textnormal{0.9}} &61.8\textsubscript{\textnormal{$\pm$}\textnormal{0.5}} &62.9\textsubscript{\textnormal{$\pm$}\textnormal{0.2}} &\textbf{\textcolor{red}{66.7\textsubscript{\textnormal{$\pm$}\textnormal{0.3}}}} &62.8\textsubscript{\textnormal{$\pm$}\textnormal{0.7}} &OOT &  \\
 & &  & 909 &0.5\% &58.3\textsubscript{\textnormal{$\pm$}\textnormal{0.0}} &60.3\textsubscript{\textnormal{$\pm$}\textnormal{0.0}} &\underline{\textcolor{blue}{67.2\textsubscript{\textnormal{$\pm$}\textnormal{0.1}}}} &\textbf{\textcolor{red}{67.4\textsubscript{\textnormal{$\pm$}\textnormal{0.3}}}} &63.9\textsubscript{\textnormal{$\pm$}\textnormal{0.1}} &64.3\textsubscript{\textnormal{$\pm$}\textnormal{0.4}} &66.9\textsubscript{\textnormal{$\pm$}\textnormal{0.3}} &62.4\textsubscript{\textnormal{$\pm$}\textnormal{0.2}} &62.6\textsubscript{\textnormal{$\pm$}\textnormal{0.2}} &62.5\textsubscript{\textnormal{$\pm$}\textnormal{0.0}} &66.4\textsubscript{\textnormal{$\pm$}\textnormal{0.3}} &63.6\textsubscript{\textnormal{$\pm$}\textnormal{0.3}} &OOT &  \\
\cmidrule{3-19}
 & & \multirow{3}{*}[-0.4ex]{GCN} & 90 &0.05\% &47.1\textsubscript{\textnormal{$\pm$}\textnormal{0.0}} &54.5\textsubscript{\textnormal{$\pm$}\textnormal{0.0}} &53.7\textsubscript{\textnormal{$\pm$}\textnormal{1.6}} &62.7\textsubscript{\textnormal{$\pm$}\textnormal{0.4}} &55.6\textsubscript{\textnormal{$\pm$}\textnormal{0.4}} &61.0\textsubscript{\textnormal{$\pm$}\textnormal{0.5}} &55.9\textsubscript{\textnormal{$\pm$}\textnormal{5.8}} &63.3\textsubscript{\textnormal{$\pm$}\textnormal{0.3}} &63.8\textsubscript{\textnormal{$\pm$}\textnormal{0.3}} &64.4\textsubscript{\textnormal{$\pm$}\textnormal{0.5}} &\textbf{\textcolor{red}{66.1\textsubscript{\textnormal{$\pm$}\textnormal{0.4}}}} &\underline{\textcolor{blue}{65.5\textsubscript{\textnormal{$\pm$}\textnormal{1.0}}}} &OOT & \multirow{3}{*}[-0.4ex]{71.1±0.0} \\
 & &  & 454 &0.25\% &56.8\textsubscript{\textnormal{$\pm$}\textnormal{0.0}} &60.3\textsubscript{\textnormal{$\pm$}\textnormal{0.0}} &64.2\textsubscript{\textnormal{$\pm$}\textnormal{0.2}} &65.4\textsubscript{\textnormal{$\pm$}\textnormal{0.4}} &61.6\textsubscript{\textnormal{$\pm$}\textnormal{0.5}} &64.7\textsubscript{\textnormal{$\pm$}\textnormal{0.2}} &63.2\textsubscript{\textnormal{$\pm$}\textnormal{0.3}} &66.4\textsubscript{\textnormal{$\pm$}\textnormal{0.1}} &66.7\textsubscript{\textnormal{$\pm$}\textnormal{0.4}} &\textbf{\textcolor{red}{67.5\textsubscript{\textnormal{$\pm$}\textnormal{0.3}}}} &\underline{\textcolor{blue}{67.2\textsubscript{\textnormal{$\pm$}\textnormal{0.4}}}} &65.8\textsubscript{\textnormal{$\pm$}\textnormal{0.4}} &OOT & \\
 & &  & 909 &0.5\% &60.3\textsubscript{\textnormal{$\pm$}\textnormal{0.0}} &62.1\textsubscript{\textnormal{$\pm$}\textnormal{0.0}} &65.1\textsubscript{\textnormal{$\pm$}\textnormal{0.4}} &66.6\textsubscript{\textnormal{$\pm$}\textnormal{0.2}} &63.4\textsubscript{\textnormal{$\pm$}\textnormal{0.4}} &65.8\textsubscript{\textnormal{$\pm$}\textnormal{0.1}} &66.8\textsubscript{\textnormal{$\pm$}\textnormal{0.3}} &67.6\textsubscript{\textnormal{$\pm$}\textnormal{0.0}} &67.6\textsubscript{\textnormal{$\pm$}\textnormal{0.3}} &\textbf{\textcolor{red}{68.0\textsubscript{\textnormal{$\pm$}\textnormal{0.3}}}} &\underline{\textcolor{blue}{67.8\textsubscript{\textnormal{$\pm$}\textnormal{0.2}}}} &66.2\textsubscript{\textnormal{$\pm$}\textnormal{0.5}} &OOT & \\
 \cmidrule{2-19}
 & \multirow{6}{*}[-1.5ex]{\rotatebox{90}{Products}} & \multirow{3}{*}[-0.4ex]{SGC} &612 &0.025\% &51.6\textsubscript{\textnormal{$\pm$}\textnormal{1.3}} &48.6\textsubscript{\textnormal{$\pm$}\textnormal{0.6}} &64.0\textsubscript{\textnormal{$\pm$}\textnormal{0.2}} &\underline{\textcolor{blue}{64.6\textsubscript{\textnormal{$\pm$}\textnormal{0.1}}}} &62.1\textsubscript{\textnormal{$\pm$}\textnormal{0.1}} &63.6\textsubscript{\textnormal{$\pm$}\textnormal{0.3}} &\textbf{\textcolor{red}{64.9\textsubscript{\textnormal{$\pm$}\textnormal{0.1}}}} &57.7\textsubscript{\textnormal{$\pm$}\textnormal{0.2}} &58.9\textsubscript{\textnormal{$\pm$}\textnormal{0.1}} &58.5\textsubscript{\textnormal{$\pm$}\textnormal{0.6}} &62.2\textsubscript{\textnormal{$\pm$}\textnormal{0.1}} &61.1\textsubscript{\textnormal{$\pm$}\textnormal{0.4}} &OOT &\multirow{3}{*}[-0.4ex]{64.7±0.1} \\
 & &  &1225 &0.05\% &55.2\textsubscript{\textnormal{$\pm$}\textnormal{0.8}} &52.2\textsubscript{\textnormal{$\pm$}\textnormal{0.7}} &\textbf{\textcolor{red}{64.0\textsubscript{\textnormal{$\pm$}\textnormal{0.1}}}} &62.4\textsubscript{\textnormal{$\pm$}\textnormal{0.1}} &61.0\textsubscript{\textnormal{$\pm$}\textnormal{0.3}} &\underline{\textcolor{blue}{62.7\textsubscript{\textnormal{$\pm$}\textnormal{0.4}}}} &62.3\textsubscript{\textnormal{$\pm$}\textnormal{0.2}} &58.2\textsubscript{\textnormal{$\pm$}\textnormal{0.3}} &61.0\textsubscript{\textnormal{$\pm$}\textnormal{0.1}} &60.9\textsubscript{\textnormal{$\pm$}\textnormal{0.1}} &62.2\textsubscript{\textnormal{$\pm$}\textnormal{0.2}} &62.4\textsubscript{\textnormal{$\pm$}\textnormal{0.2}} &OOT & \\
 & &  &2449 &0.1\% &58.0\textsubscript{\textnormal{$\pm$}\textnormal{0.8}} &55.4\textsubscript{\textnormal{$\pm$}\textnormal{0.4}} &\textbf{\textcolor{red}{64.4\textsubscript{\textnormal{$\pm$}\textnormal{0.4}}}} &62.8\textsubscript{\textnormal{$\pm$}\textnormal{0.1}} &61.4\textsubscript{\textnormal{$\pm$}\textnormal{0.2}} &62.3\textsubscript{\textnormal{$\pm$}\textnormal{0.2}} &\underline{\textcolor{blue}{64.3\textsubscript{\textnormal{$\pm$}\textnormal{0.3}}}} &60.8\textsubscript{\textnormal{$\pm$}\textnormal{0.2}} &61.3\textsubscript{\textnormal{$\pm$}\textnormal{0.1}} &61.3\textsubscript{\textnormal{$\pm$}\textnormal{0.2}} &61.9\textsubscript{\textnormal{$\pm$}\textnormal{0.2}} &63.1\textsubscript{\textnormal{$\pm$}\textnormal{0.2}} &OOT & \\
\cmidrule{3-19}
 & & \multirow{3}{*}[-0.2ex]{GCN} &612 &0.025\% &57.8\textsubscript{\textnormal{$\pm$}\textnormal{0.5}} &55.4\textsubscript{\textnormal{$\pm$}\textnormal{0.8}} &63.7\textsubscript{\textnormal{$\pm$}\textnormal{0.3}} &65.5\textsubscript{\textnormal{$\pm$}\textnormal{0.2}} &61.2\textsubscript{\textnormal{$\pm$}\textnormal{0.3}} &62.2\textsubscript{\textnormal{$\pm$}\textnormal{0.6}} &64.0\textsubscript{\textnormal{$\pm$}\textnormal{0.4}} &66.5\textsubscript{\textnormal{$\pm$}\textnormal{0.1}}  &\underline{\textcolor{blue}{68.0\textsubscript{\textnormal{$\pm$}\textnormal{0.3}}}} &66.2\textsubscript{\textnormal{$\pm$}\textnormal{0.1}}
 &67.1\textsubscript{\textnormal{$\pm$}\textnormal{0.2}} &\textbf{\textcolor{red}{68.5\textsubscript{\textnormal{$\pm$}\textnormal{0.3}}}} &OOT &\multirow{3}{*}[-0.4ex]{73.1±0.1} \\
 & &  &1225 &0.05\% &61.6\textsubscript{\textnormal{$\pm$}\textnormal{0.5}} &57.6\textsubscript{\textnormal{$\pm$}\textnormal{0.7}} &67.0\textsubscript{\textnormal{$\pm$}\textnormal{0.2}} &66.6\textsubscript{\textnormal{$\pm$}\textnormal{0.3}} &62.6\textsubscript{\textnormal{$\pm$}\textnormal{0.4}} &64.2\textsubscript{\textnormal{$\pm$}\textnormal{0.2}} &65.9\textsubscript{\textnormal{$\pm$}\textnormal{0.2}} &68.4\textsubscript{\textnormal{$\pm$}\textnormal{0.4}} &\underline{\textcolor{blue}{68.8\textsubscript{\textnormal{$\pm$}\textnormal{0.1}}}} &68.5\textsubscript{\textnormal{$\pm$}\textnormal{0.2}} &67.9\textsubscript{\textnormal{$\pm$}\textnormal{0.3}} &\textbf{\textcolor{red}{69.8\textsubscript{\textnormal{$\pm$}\textnormal{0.3}}}} &OOT & \\
 & &  &2449 &0.1\%  &65.3\textsubscript{\textnormal{$\pm$}\textnormal{0.5}} &59.1\textsubscript{\textnormal{$\pm$}\textnormal{0.5}} &68.0\textsubscript{\textnormal{$\pm$}\textnormal{0.2}} &68.3\textsubscript{\textnormal{$\pm$}\textnormal{0.2}} &65.8\textsubscript{\textnormal{$\pm$}\textnormal{0.2}} &66.8\textsubscript{\textnormal{$\pm$}\textnormal{0.1}} &66.1\textsubscript{\textnormal{$\pm$}\textnormal{0.3}} &68.4\textsubscript{\textnormal{$\pm$}\textnormal{0.3}} &70.0\textsubscript{\textnormal{$\pm$}\textnormal{0.1}} &69.8\textsubscript{\textnormal{$\pm$}\textnormal{0.2}} &\underline{\textcolor{blue}{70.1\textsubscript{\textnormal{$\pm$}\textnormal{0.3}}}} &\textbf{\textcolor{red}{71.1\textsubscript{\textnormal{$\pm$}\textnormal{0.3}}}} &OOT & \\
\midrule
\midrule
\multirow{12}{*}[-0.5cm]{\rotatebox{90}{Inductive}} & \multirow{6}{*}[-1.5ex]{\rotatebox{90}{Flickr}} & \multirow{3}{*}[-0.4ex]{SGC} & 44 &0.1\% &27.6\textsubscript{\textnormal{$\pm$}\textnormal{0.1}} &34.5\textsubscript{\textnormal{$\pm$}\textnormal{0.1}} &43.7\textsubscript{\textnormal{$\pm$}\textnormal{0.5}} &43.7\textsubscript{\textnormal{$\pm$}\textnormal{0.2}} &41.8\textsubscript{\textnormal{$\pm$}\textnormal{0.4}} &42.2\textsubscript{\textnormal{$\pm$}\textnormal{0.1}} &43.6\textsubscript{\textnormal{$\pm$}\textnormal{0.3}} &40.3\textsubscript{\textnormal{$\pm$}\textnormal{0.0}} &37.5\textsubscript{\textnormal{$\pm$}\textnormal{0.2}} &\textbf{\textcolor{red}{45.3\textsubscript{\textnormal{$\pm$}\textnormal{0.2}}}} &\underline{\textcolor{blue}{45.3\textsubscript{\textnormal{$\pm$}\textnormal{0.7}}}} &33.6\textsubscript{\textnormal{$\pm$}\textnormal{0.4}} &OOT & \multirow{3}{*}[-0.4ex]{44.2±0.0} \\
 & &  & 223 &0.5\% &33.5\textsubscript{\textnormal{$\pm$}\textnormal{0.0}} &36.1\textsubscript{\textnormal{$\pm$}\textnormal{0.0}} &42.2\textsubscript{\textnormal{$\pm$}\textnormal{0.2}} &43.8\textsubscript{\textnormal{$\pm$}\textnormal{0.6}} &42.6\textsubscript{\textnormal{$\pm$}\textnormal{0.2}} &\underline{\textcolor{blue}{43.9\textsubscript{\textnormal{$\pm$}\textnormal{0.1}}}} &41.6\textsubscript{\textnormal{$\pm$}\textnormal{1.6}} &40.8\textsubscript{\textnormal{$\pm$}\textnormal{0.1}} &36.2\textsubscript{\textnormal{$\pm$}\textnormal{0.2}} &43.6\textsubscript{\textnormal{$\pm$}\textnormal{0.1}} &\textbf{\textcolor{red}{45.7\textsubscript{\textnormal{$\pm$}\textnormal{0.4}}}} &37.4\textsubscript{\textnormal{$\pm$}\textnormal{0.2}} &OOT &  \\
 & &  & 446 &1\% &34.4\textsubscript{\textnormal{$\pm$}\textnormal{0.0}} &36.5\textsubscript{\textnormal{$\pm$}\textnormal{0.0}} &41.1\textsubscript{\textnormal{$\pm$}\textnormal{0.8}} &\underline{\textcolor{blue}{43.7\textsubscript{\textnormal{$\pm$}\textnormal{0.3}}}} &42.1\textsubscript{\textnormal{$\pm$}\textnormal{0.4}} &43.3\textsubscript{\textnormal{$\pm$}\textnormal{0.1}} &43.2\textsubscript{\textnormal{$\pm$}\textnormal{0.4}} &42.7\textsubscript{\textnormal{$\pm$}\textnormal{0.4}} &34.3\textsubscript{\textnormal{$\pm$}\textnormal{0.3}} &42.6\textsubscript{\textnormal{$\pm$}\textnormal{0.1}} &\textbf{\textcolor{red}{46.1\textsubscript{\textnormal{$\pm$}\textnormal{0.5}}}} &38.1\textsubscript{\textnormal{$\pm$}\textnormal{0.2}} &OOT &  \\
\cmidrule{3-19}
 & & \multirow{3}{*}[-0.6ex]{GCN} & 44 &0.1\% &39.2\textsubscript{\textnormal{$\pm$}\textnormal{0.0}} &40.7\textsubscript{\textnormal{$\pm$}\textnormal{0.0}} &43.3\textsubscript{\textnormal{$\pm$}\textnormal{0.3}} &\underline{\textcolor{blue}{45.4\textsubscript{\textnormal{$\pm$}\textnormal{0.3}}}} &43.3\textsubscript{\textnormal{$\pm$}\textnormal{1.1}} &44.9\textsubscript{\textnormal{$\pm$}\textnormal{0.9}} &42.4\textsubscript{\textnormal{$\pm$}\textnormal{0.2}} &44.5\textsubscript{\textnormal{$\pm$}\textnormal{0.4}} &44.8\textsubscript{\textnormal{$\pm$}\textnormal{0.5}} &45.2\textsubscript{\textnormal{$\pm$}\textnormal{0.3}} &\textbf{\textcolor{red}{46.9\textsubscript{\textnormal{$\pm$}\textnormal{0.3}}}} &44.6\textsubscript{\textnormal{$\pm$}\textnormal{0.5}} &OOT &\multirow{3}{*}[-0.4ex]{46.8±0.2} \\
 & &  & 223 &0.5\% &42.4\textsubscript{\textnormal{$\pm$}\textnormal{0.0}} &41.4\textsubscript{\textnormal{$\pm$}\textnormal{0.0}} &44.6\textsubscript{\textnormal{$\pm$}\textnormal{0.4}} &45.8\textsubscript{\textnormal{$\pm$}\textnormal{0.3}} &43.5\textsubscript{\textnormal{$\pm$}\textnormal{1.0}} &45.0\textsubscript{\textnormal{$\pm$}\textnormal{0.2}} &44.9\textsubscript{\textnormal{$\pm$}\textnormal{0.3}} &45.0\textsubscript{\textnormal{$\pm$}\textnormal{0.2}} &\underline{\textcolor{blue}{46.4\textsubscript{\textnormal{$\pm$}\textnormal{0.2}}}} &45.6\textsubscript{\textnormal{$\pm$}\textnormal{0.7}} &\textbf{\textcolor{red}{47.0\textsubscript{\textnormal{$\pm$}\textnormal{0.1}}}} &45.2\textsubscript{\textnormal{$\pm$}\textnormal{0.9}} &OOT & \\
 & &  & 446 &1\% &43.4\textsubscript{\textnormal{$\pm$}\textnormal{0.0}} &41.4\textsubscript{\textnormal{$\pm$}\textnormal{0.0}} &44.4\textsubscript{\textnormal{$\pm$}\textnormal{0.1}} &45.7\textsubscript{\textnormal{$\pm$}\textnormal{0.2}} &42.8\textsubscript{\textnormal{$\pm$}\textnormal{0.2}} &44.8\textsubscript{\textnormal{$\pm$}\textnormal{0.4}} &45.2\textsubscript{\textnormal{$\pm$}\textnormal{0.2}} &44.6\textsubscript{\textnormal{$\pm$}\textnormal{0.3}} &\underline{\textcolor{blue}{46.7\textsubscript{\textnormal{$\pm$}\textnormal{0.2}}}} &44.9\textsubscript{\textnormal{$\pm$}\textnormal{0.1}} &\textbf{\textcolor{red}{47.2\textsubscript{\textnormal{$\pm$}\textnormal{0.1}}}} &45.5\textsubscript{\textnormal{$\pm$}\textnormal{0.1}} &OOT & \\
\cmidrule{2-19}
& \multirow{6}{*}[-1.5ex]{\rotatebox{90}{Reddit}} & \multirow{3}{*}[-0.4ex]{SGC} & 153 &0.05\% &55.8\textsubscript{\textnormal{$\pm$}\textnormal{0.2}} &54.0\textsubscript{\textnormal{$\pm$}\textnormal{0.1}} &89.7\textsubscript{\textnormal{$\pm$}\textnormal{0.6}} &91.1\textsubscript{\textnormal{$\pm$}\textnormal{0.1}} &90.8\textsubscript{\textnormal{$\pm$}\textnormal{0.1}} &91.0\textsubscript{\textnormal{$\pm$}\textnormal{0.0}} &90.5\textsubscript{\textnormal{$\pm$}\textnormal{0.3}} &90.3\textsubscript{\textnormal{$\pm$}\textnormal{0.8}} &\underline{\textcolor{blue}{92.0\textsubscript{\textnormal{$\pm$}\textnormal{0.0}}}} &\textbf{\textcolor{red}{92.1\textsubscript{\textnormal{$\pm$}\textnormal{0.0}}}} &90.9\textsubscript{\textnormal{$\pm$}\textnormal{0.2}} &59.4\textsubscript{\textnormal{$\pm$}\textnormal{1.5}} &OOT & \multirow{3}{*}[-0.4ex]{93.2±0.0} \\
 & &  & 769 &0.1\% &74.1\textsubscript{\textnormal{$\pm$}\textnormal{0.1}} &78.6\textsubscript{\textnormal{$\pm$}\textnormal{0.0}} &91.8\textsubscript{\textnormal{$\pm$}\textnormal{0.2}} &90.4\textsubscript{\textnormal{$\pm$}\textnormal{0.1}} &91.5\textsubscript{\textnormal{$\pm$}\textnormal{0.0}} &91.8\textsubscript{\textnormal{$\pm$}\textnormal{0.0}} &\underline{\textcolor{blue}{91.9\textsubscript{\textnormal{$\pm$}\textnormal{0.0}}}} &88.1\textsubscript{\textnormal{$\pm$}\textnormal{2.8}} &90.9\textsubscript{\textnormal{$\pm$}\textnormal{0.1}} &91.0\textsubscript{\textnormal{$\pm$}\textnormal{0.0}} &\textbf{\textcolor{red}{92.6\textsubscript{\textnormal{$\pm$}\textnormal{0.2}}}} &81.7\textsubscript{\textnormal{$\pm$}\textnormal{0.7}} &OOT &  \\
 & &  & 1539 &0.2\% &83.3\textsubscript{\textnormal{$\pm$}\textnormal{0.0}} &83.8\textsubscript{\textnormal{$\pm$}\textnormal{0.0}} &\underline{\textcolor{blue}{92.1\textsubscript{\textnormal{$\pm$}\textnormal{0.3}}}} &90.4\textsubscript{\textnormal{$\pm$}\textnormal{0.1}} &92.0\textsubscript{\textnormal{$\pm$}\textnormal{0.2}} &92.1\textsubscript{\textnormal{$\pm$}\textnormal{0.0}} &86.3\textsubscript{\textnormal{$\pm$}\textnormal{5.6}} &91.7\textsubscript{\textnormal{$\pm$}\textnormal{0.2}} &90.0\textsubscript{\textnormal{$\pm$}\textnormal{0.0}} &89.6\textsubscript{\textnormal{$\pm$}\textnormal{0.1}} &\textbf{\textcolor{red}{92.6\textsubscript{\textnormal{$\pm$}\textnormal{0.3}}}} &86.7\textsubscript{\textnormal{$\pm$}\textnormal{0.1}} &OOT &  \\
\cmidrule{3-19}
 & & \multirow{3}{*}[-0.6ex]{GCN} & 153 &0.05\% &60.4\textsubscript{\textnormal{$\pm$}\textnormal{0.0}} &58.6\textsubscript{\textnormal{$\pm$}\textnormal{0.1}} &56.8\textsubscript{\textnormal{$\pm$}\textnormal{2.1}} &85.8\textsubscript{\textnormal{$\pm$}\textnormal{0.1}} &75.8\textsubscript{\textnormal{$\pm$}\textnormal{1.9}} &88.6\textsubscript{\textnormal{$\pm$}\textnormal{0.3}} &72.9\textsubscript{\textnormal{$\pm$}\textnormal{4.9}} &88.9\textsubscript{\textnormal{$\pm$}\textnormal{1.2}} &91.2\textsubscript{\textnormal{$\pm$}\textnormal{0.1}} &\textbf{\textcolor{red}{91.3\textsubscript{\textnormal{$\pm$}\textnormal{0.1}}}} &89.2\textsubscript{\textnormal{$\pm$}\textnormal{0.5}} &\underline{\textcolor{blue}{90.0\textsubscript{\textnormal{$\pm$}\textnormal{0.5}}}} &OOT &\multirow{3}{*}[-0.4ex]{94.2±0.0} \\
 & &  & 769 &0.1\% &81.7\textsubscript{\textnormal{$\pm$}\textnormal{0.0}} &81.7\textsubscript{\textnormal{$\pm$}\textnormal{0.0}} &87.4\textsubscript{\textnormal{$\pm$}\textnormal{0.4}} &90.5\textsubscript{\textnormal{$\pm$}\textnormal{0.2}} &87.9\textsubscript{\textnormal{$\pm$}\textnormal{1.1}} &91.2\textsubscript{\textnormal{$\pm$}\textnormal{0.2}} &89.6\textsubscript{\textnormal{$\pm$}\textnormal{2.5}} &91.8\textsubscript{\textnormal{$\pm$}\textnormal{0.3}} &\underline{\textcolor{blue}{92.4\textsubscript{\textnormal{$\pm$}\textnormal{0.0}}}} &\textbf{\textcolor{red}{92.6\textsubscript{\textnormal{$\pm$}\textnormal{0.2}}}} &90.9\textsubscript{\textnormal{$\pm$}\textnormal{0.3}} &89.4\textsubscript{\textnormal{$\pm$}\textnormal{0.5}} &OOT & \\
 & &  & 1539 &0.2\% &87.1\textsubscript{\textnormal{$\pm$}\textnormal{0.0}} &86.9\textsubscript{\textnormal{$\pm$}\textnormal{0.0}} &91.4\textsubscript{\textnormal{$\pm$}\textnormal{0.4}} &89.5\textsubscript{\textnormal{$\pm$}\textnormal{0.3}} &88.2\textsubscript{\textnormal{$\pm$}\textnormal{2.1}} &91.9\textsubscript{\textnormal{$\pm$}\textnormal{0.1}} &91.2\textsubscript{\textnormal{$\pm$}\textnormal{0.3}} &92.2\textsubscript{\textnormal{$\pm$}\textnormal{0.1}} &\textbf{\textcolor{red}{92.7\textsubscript{\textnormal{$\pm$}\textnormal{0.1}}}} &\underline{\textcolor{blue}{92.5\textsubscript{\textnormal{$\pm$}\textnormal{0.3}}}} &92.4\textsubscript{\textnormal{$\pm$}\textnormal{0.1}} &91.2\textsubscript{\textnormal{$\pm$}\textnormal{0.1}} &OOT & \\
\bottomrule
\end{tabular}

%% file: tables/init.tex
\begin{tabular}{llccc}
    \toprule
    Label & Feature & GCond & DosCond & DM \\
    \midrule
    \multirow{3}{*}{Balanced} & Random Noise & 57.4 & 56.1 & 62.4 \\
                              & Random Subgraph & 56.2 & 55.2 & 62.9 \\
                              & k-Center & 57.6 & 57.2 & 62.7 \\
    \midrule
    \multirow{3}{*}{Proportional} & Random Noise & 58.4 & 60.4 & 64.5 \\
                              & Random Subgraph & 58.4 & 59.4 & 64.1 \\
                              & k-Center & 57.9 & 59.3 & 63.7 \\
    \bottomrule
\end{tabular}

%% file: tables/cora-35-evaluator-test-acc.tex
\begin{tabular}{c|cccccccccc}
\toprule
 \textbf{Validator} & GCond & GCondX & DosCond & DosCondX & SGDD & GCDM &GCDMX & DM & SFGC & Avg. \\
\midrule\midrule
\textbf{GCN} & 73.8 (1.0x) & 80.6 (1.0x) & 78.3 (1.0x) & 80.7 (1.0x) & 78.7 (1.0x) & 79.4 (1.0x) &78.9 (1.0x) & 80.2 (1.0x) & 79.6 (1.0x) & 78.9 (1.0x) \\
\textbf{SGC} & 67.6 (0.5x) & 80.7 (0.7x) & 78.2 (0.8x) & 81.1 (0.9x) & 64.2 (0.7x) & 80.0 (1.1x) &78.5 (1.0x) & 80.4 (1.0x) & 79.8 (1.0x) & 76.7 (0.9x) \\
\textbf{GNTK} & 51.8 (0.2x) & 79.7 (0.2x) & 78.7 (0.8x) & 78.8 (0.6x) & 73.3 (0.6x) & 57.6 (1.1x) &79.5 (0.6x) & 78.0 (0.2x) & 79.3 (0.9x) & 73.0 (0.6x) \\
\bottomrule
\end{tabular}

%% file: sections/related_work.tex
\section{Related Work}
\textbf{Graph condensation} methods can be categorised into the following branches. (1) Gradient matching methods align the gradients of GNN models trained on the original and the synthetic graphs, including GCond~\cite{jinGraphCondensationGraph2022a}, DosCond~\cite{jinCondensingGraphsOneStep2022a}, and SGDD~\cite{yangDoesGraphDistillation2023}. (2) Distribution matching methods align the embeddings of the original and the synthetic graphs, such as GCDM~\cite{liuGraphCondensationReceptive2022}, CaT~\cite{liuCaTBalancedContinual2023a}, and PUMA~\cite{liuPUMAEfficientContinual2023}. (3)~Trajectory matching methods align the parameters of GNN models trained on the original and the synthetic graphs as in SFGC~\cite{zhengStructurefreeGraphCondensation2023a} and GEOM~\cite{geom}. (4) Kernel matching methods rely on graph neural tangent kernel to align the original and the synthetic graphs, including GC-SNTK~\cite{gc-sntk} and KIDD~\cite{xuKernelRidgeRegressionBased2023a}. (5) Eigen matching methods focus on matching the eigen-decomposition features of the structure of the original and the synthetic graphs as in SGDD~\cite{yangDoesGraphDistillation2023} and GDEM~\cite{liuGraphCondensationEigenbasis2023}.
Note that there are several computer vision dataset condensation methods mainly focused on image data~\cite{cazenavetteDatasetDistillationMatching2022, wangCAFELearningCondense2022, wangDatasetDistillation2018, zhaoDatasetCondensationDifferentiable2021, zhaoDatasetCondensationDistribution2023a, zhaoDatasetCondensationGradient2021a}, as well as a corresponding benchmark DC-Bench~\cite{cuiDCBENCHDatasetCondensation2022}. However, a direct extension to graph representation learning is nontrivial since graph data involves not only features but also topological structures with interdependence.

\textbf{Tasks and applications of graph condensation} of existing work mainly focus on node classification.
Recently, graph classification~\cite{xuKernelRidgeRegressionBased2023a}, link prediction~\cite{yangDoesGraphDistillation2023}, fairness~\cite{fengFairGraphDistillation2023}, and heterogeneous graphs have gradually attracted attention from the research community.
Additionally, graph condensation has served as a powerful tool for various downstream tasks, including continual graph learning~\cite{liuCaTBalancedContinual2023a,liuPUMAEfficientContinual2023}, inductive inference~\cite{gaoGraphCondensationInductive2023}, federated learning~\cite{panFedGKDUnleashingPower2023}, neural architecture search~\cite{dingFasterHyperparameterSearch2022}, etc.
%
%
%
%
%

%% file: sections/conclusion_future.tex
\section{Conclusion and Future Work}
\label{sec:conclusion}
\textbf{Conclusion}: This paper introduces a large-scale graph condensation benchmark, \texttt{GCondenser}, which establishes a standardised GC paradigm to evaluate and compare mainstream GC methods. This benchmark highlights the differences among various GC components and methods. Moreover, a comprehensive experimental study is conducted, demonstrating that \texttt{GCondenser} improve the condensation quality of existing methods through extensive hyperparameter tuning and appropriate validation approaches.

\textbf{Societal impact}: \texttt{GCondenser} provides a unified graph GC paradigm and various GC methods, making the implementation of graph condensation techniques easier. On the positive side, it enhances the development of GC research. On the slightly negative side, the improper use of GC can impact downstream tasks, such as data sensitivity and the robustness of the training model.

\textbf{Limitation and future work}: The current state of \texttt{GCondenser} reflects the development of graph condensation methods on datasets with a relatively small scale. In the near future, \texttt{GCondenser} will be expanded to support more emerging GC methods and additional large-scale graph datasets, such Papers100M~\cite{huOpenGraphBenchmark2020}. Further, \texttt{GCondenser} will include a wider collection of downstream applications of graph condensation, such as graph neural architecture search as in GraphGym~\cite{youDesignSpaceGraph2020}, link prediction and graph classification. The straightforward extension to new methods, tasks and datasets will ensure the \texttt{GCondenser} to be an updated and powerful benchmark for graph condensation.

%% file: sections/appendix.tex
\appendix
\section{Implementation Details}
\subsection{Graph Condensation Settings}\label{app:setting}
\begin{figure}[!h]
    \centering
    \includegraphics[width=1\textwidth]{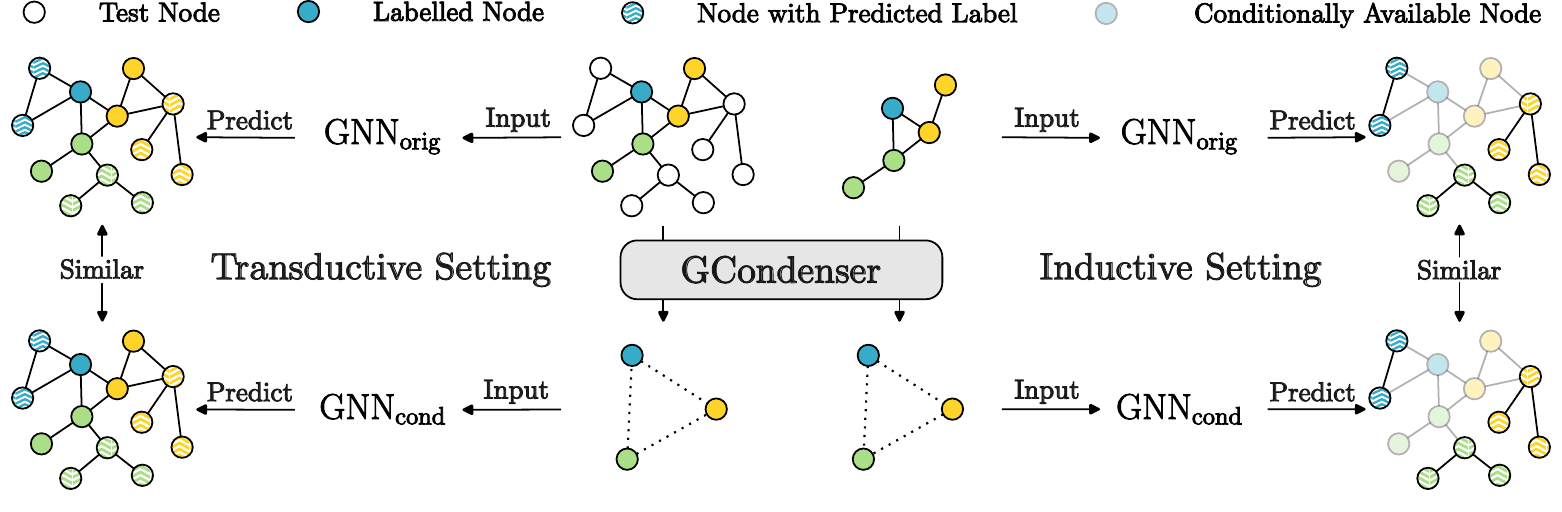}
    \caption{\texttt{GCondenser} for graph condensation with transductive (left) and inductive (right) settings.}
    \label{fig:setting}
\end{figure}

Transductive and inductive are two settings for graph dataset condensation, as introduced by~\cite{jinGraphCondensationGraph2022a}. The differences between these settings are illustrated in Figure~\ref{fig:setting}. In the transductive setting, test nodes and their induced edges are available during training, whereas in the inductive setting, these nodes and edges are unavailable. In the testing phase of the inductive setting, the availability of labelled nodes and their induced edges is conditional. This paper assumes that labelled nodes are unavailable during the test phase in the inductive setting.

\subsection{Label Distribution}\label{app:label_distribution}
\begin{figure}[!h]
    \centering
    \includegraphics[width=1\textwidth]{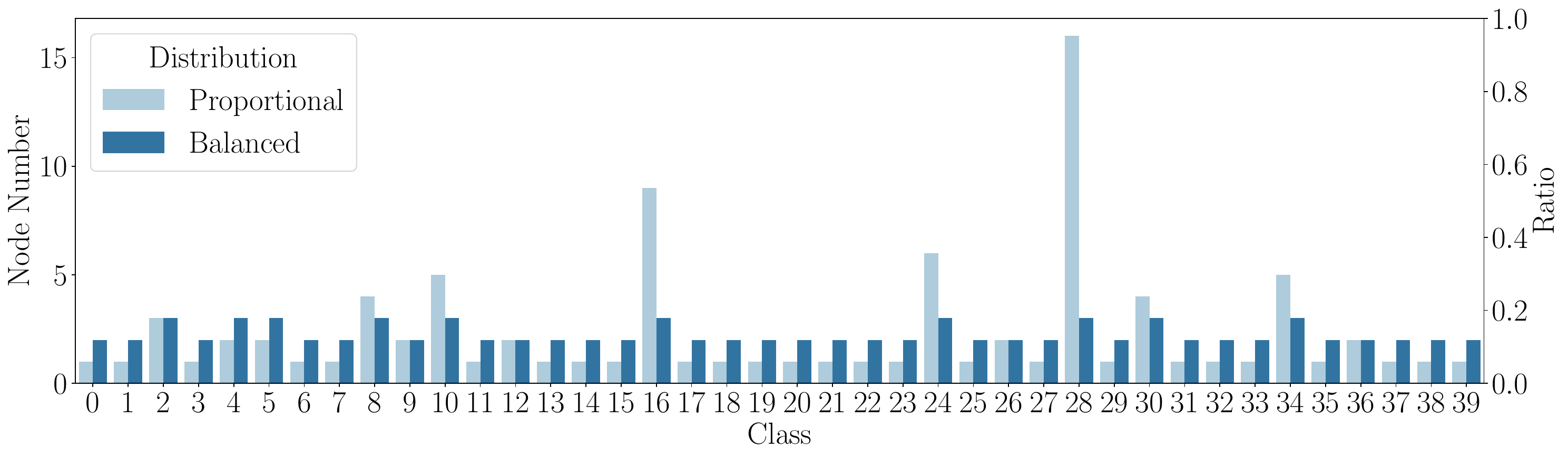}
    \caption{Comparison of the class distribution between the original and balanced label initialisation strategies with respect to the number of nodes and the class ratio in each class, using a 90-node condensed Arxiv dataset as an example.}
    \label{fig:label_distribution}
\end{figure}

\texttt{GCondenser} provides two label distribution strategies: original and balanced. The original label distribution ensures that the label distribution of the synthetic graph closely matches that of the original graph, while the balanced label distribution assigns nodes uniformly to each class. Figure~\ref{fig:label_distribution} visualises different label distributions of condensed Arxiv dataset with a 90-node budget. 

\subsection{Reproducibility}\label{app:rep}
\textbf{Condensation}: Each epoch starts with the initialisation of a backbone model within a nested loop structure. The outer loop updates the condensed graph. Subsequently, the process progresses to the inner loop, where the backbone model is continuously trained using the updated condensed graph. If a structure generator (such as GCond, DosCond, SGDD, or GCDM) is employed, it is initially optimised over several epochs, followed by a few epochs dedicated to feature updates. This iterative pattern of alternating between structural and feature updates continues. For SFGC, the backbone training follows an offline style, thereby removing the need for a nested loop. Based on the condensation framework and the methodologies of various methods, hyperparameter search spaces are clearly predefined in Tables~\ref{tab:hp1} and ~\ref{tab:hp2}.

\textbf{Validation}: The validation phase employs the same model architecture as used in the condensation process (e.g., GCN or SGC). Every 10 epochs, a validator is trained from scratch using the condensed graph and assessed against the validation set of the original graph. After 200-epoch training and validating, the best validation accuracy achieved is recorded as the validation score. The condensation process is terminated early if no improvement in validation scores is observed within five validation steps (equivalent to 50 condensation epochs); otherwise, the process continues until reaching a total of 1000 epochs.

\begin{figure}[!t]
    \centering
    \begin{subfigure}{0.45\textwidth}
        \centering
        \includegraphics[width=\linewidth]{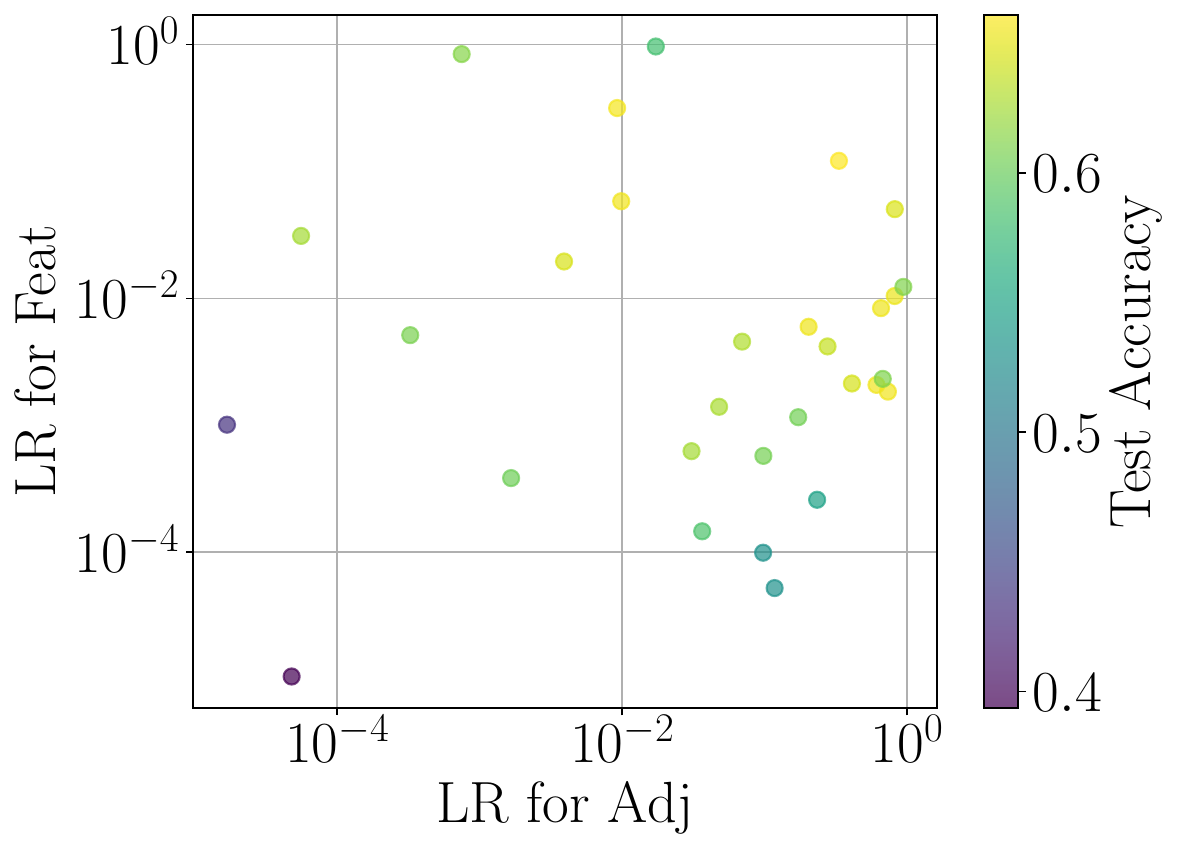}
        \caption{GCond-SGC for Arxiv-90}
    \end{subfigure}
    \hfill
    \begin{subfigure}{0.45\textwidth}
        \centering
        \includegraphics[width=\linewidth]{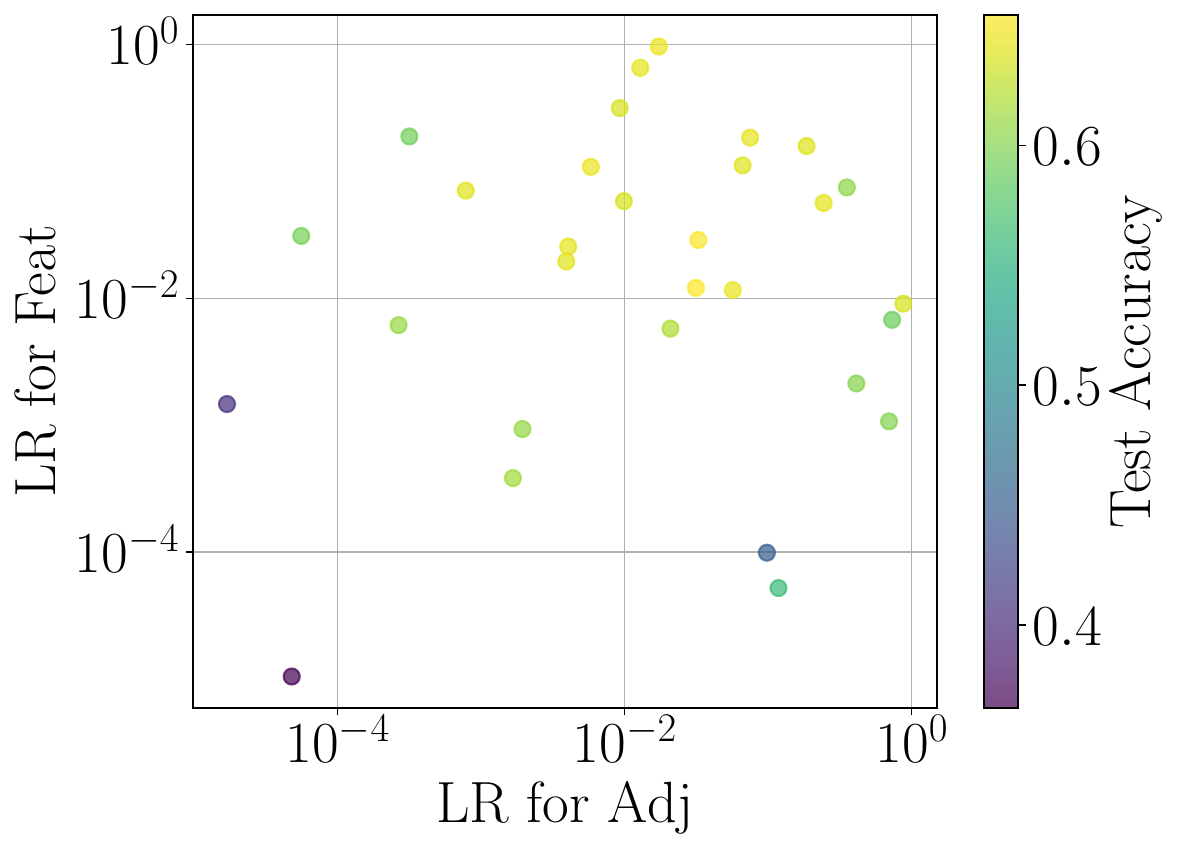}
        \caption{SGDD-SGC for Arxiv-90}
    \end{subfigure}
    \caption{Examples for overall experiments with Bayes hyperparameter sampler.}
    \label{fig:bayes}
\end{figure}

\textbf{Overall test}: After condensation, the optimal condensed graphs are loaded for testing. The test model remains the same as the backbone model:
\begin{itemize}
    \item GCN: Number of layers: 2; Hidden dimension: 256; Dropout rate: 0.5
    \item SGC: Number of the message passing hop (k): 2
\end{itemize}
The models are trained using the condensed graphs until convergence, employing an Adam optimiser with a weight decay of 0.5 and a learning rate of 0.01. The Bayesian hyperparameter sampler quickly identifies the optimal range for the hyperparameter combination, as illustrated in Figure~\ref{fig:bayes}.

\begin{table}[ht]
\centering
\tiny
\caption{Predefined hyperparameter search spaces for gradient and distribution matching methods.}
\label{tab:hp1}
\resizebox{\linewidth}{!}{
    \begin{tabular}{l|c|c|c|c}\toprule
    &GCond, SGDD, GCDM &GCondX, DosCondX &DosCond &DosCondX, DM \\\midrule
    lr for adj &log\_uniform(1e-6,1.0) &NaN &log\_uniform(1e-6,1.0) &NaN \\
    lr for feat &log\_uniform(1e-6,1.0) &log\_uniform(1e-6,1.0) &log\_uniform(1e-6,1.0) &log\_uniform(1e-6,1.0) \\
    outer loop &5 &5 &1 &1 \\
    inner loop &10 &10 &0 &0 \\
    adj update steps &10 &NaN &10 &NaN \\
    feat update steps &20 &NaN &20 &NaN \\
    \bottomrule
    \end{tabular}
}
\end{table}

\begin{table}[ht]
\centering
\tiny
\caption{Predefined hyperparameter search spaces for 
SFGC.}
\label{tab:hp2}
\resizebox{\linewidth}{!}{
    \begin{tabular}{c|c|c|c|c}\toprule
    lr for feat &lr for student model &target epochs &warm-up epochs &student epochs \\\midrule
    log\_uniform(1e-6,1.0) &log\_uniform(1e-3,1.0) &start:0, end:800, step:10 &start:0, end:100, step:10 &start:0, end:300, step:10 \\
    \bottomrule
    \end{tabular}
}
\end{table}

\subsection{Backbone Models}\label{app:backbones}
Besides GCN and SGC, which are used as condensation backbones and validators, the cross-architecture experiment also employs other backbone models, including MLP, GAT, ChebNet, SAGE, and APPNP:
\begin{itemize}
    \item MLP: Number of layers: 2; Hidden dimension: 256; Dropout rate: 0.5
    \item GAT: Number of layers: 2; Hidden dimension: 256; Number of attention heads:8; Dropout rate: 0.5; Graph sparse threshold: 0.5
    \item ChebNet: Number of layers: 2; Hidden dimension: 256; Dropout rate: 0.5; Graph sparse threshold: 0.5
    \item SAGE: Number of layers: 2; Hidden dimension: 256; Dropout rate: 0.5; Aggregator type: mean; Graph sparse threshold: 0.5
    \item APPNP: Number of layers: 2; Hidden dimension: 256; Dropout rate: 0.5; Number of iterations (propagation): 10; Teleport probability (propagation): 0.1
\end{itemize}

\subsection{Continual Graph Learning}\label{app:cgl}
Unlike the typical node classification problem, hyperparameter search in the CGL setting is challenging due to potential latency issues. This could explain why methods with fewer hyperparameters, such as DM, tend to perform better in the CGL setting. The hyperparameter choices of selected methods are shown in Table~\ref{tab:cgl-hparams}.

\begin{table}[ht]
\centering
\tiny
\caption{Hyperparameter choices for CGL experiment.}
\label{tab:cgl-hparams}
\resizebox{\linewidth}{!}{
    \begin{tabular}{l|c|c|c}\toprule
    &GCond &DosCond &DM \\\midrule
    CiteSeer &GCond: lr for adj: 1e-5; lr for feat: 1e-5; outer loop: 10; inner loop: 1 &lr for adj: 1e-5; lr for feat: 1e-5; outer loop: 1 &lr for feat: 1e-4; outer loop: 1 \\\midrule
    Arxiv &lr for adj: 1e-2; lr for feat: 1e-2; outer loop: 10; inner loop: 1 &lr for adj: 1e-2; lr for feat: 1e-2; outer loop: 1 &lr for feat: 1e-3; outer loop: 1 \\
    \bottomrule
    \end{tabular}
}
\end{table}

\section{More Experiments}
In this section, more experiments are provided in addition to the ones in the main text, including the condensation efficiency and the cross-architecture transferability.
\subsection{Condensation Efficiency}
\label{sec:efficiency-app}
\begin{figure}[!h]
    \centering
    \includegraphics[width=\linewidth]{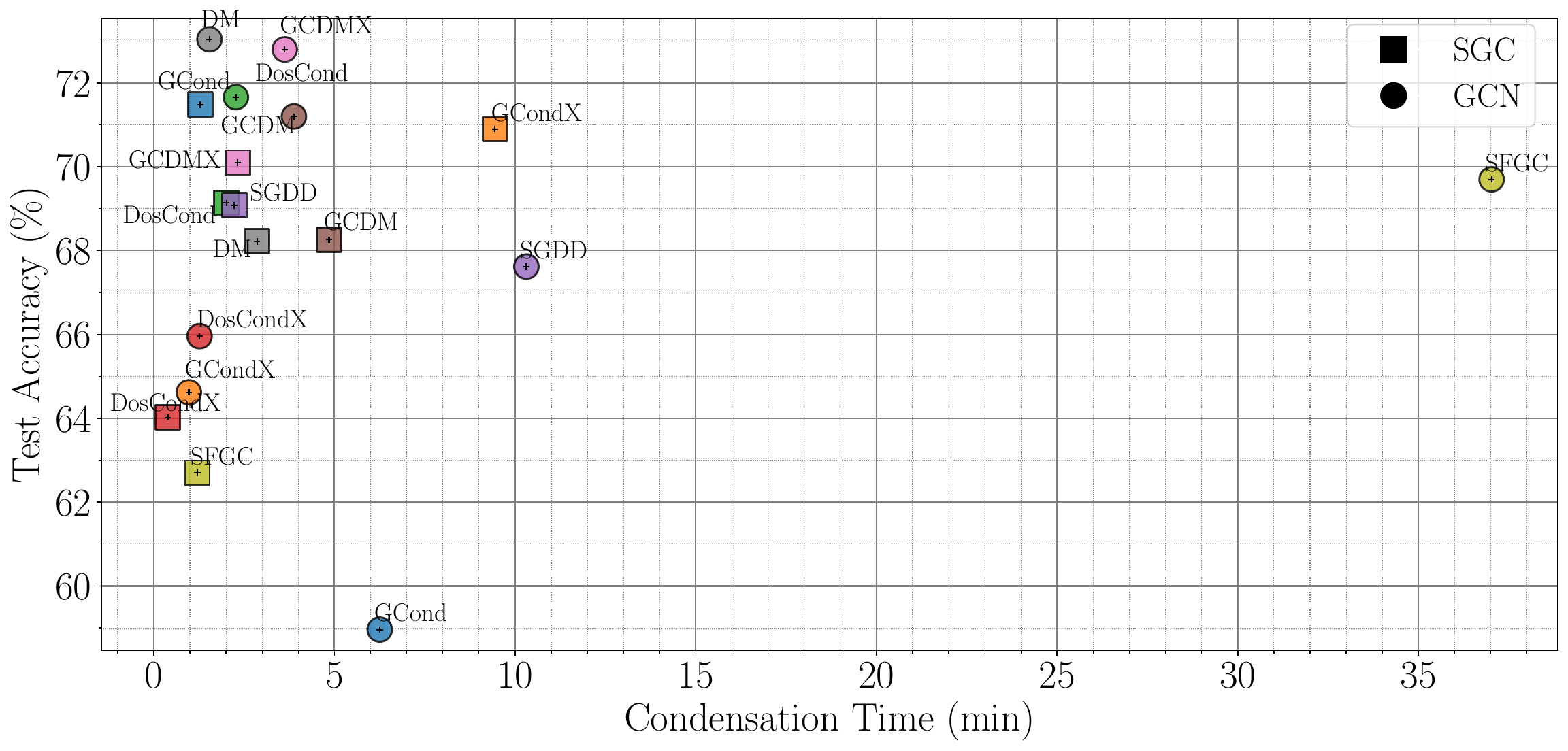}
    \caption{Test accuracy against condensation time for different GC methods on a 30-node condensed graph from the CiteSeer dataset, with backbone models GCN and SGC.}
    \label{fig:efficiency-citeseer}
\end{figure}

\begin{figure}[!h]
    \centering
    \includegraphics[width=\linewidth]{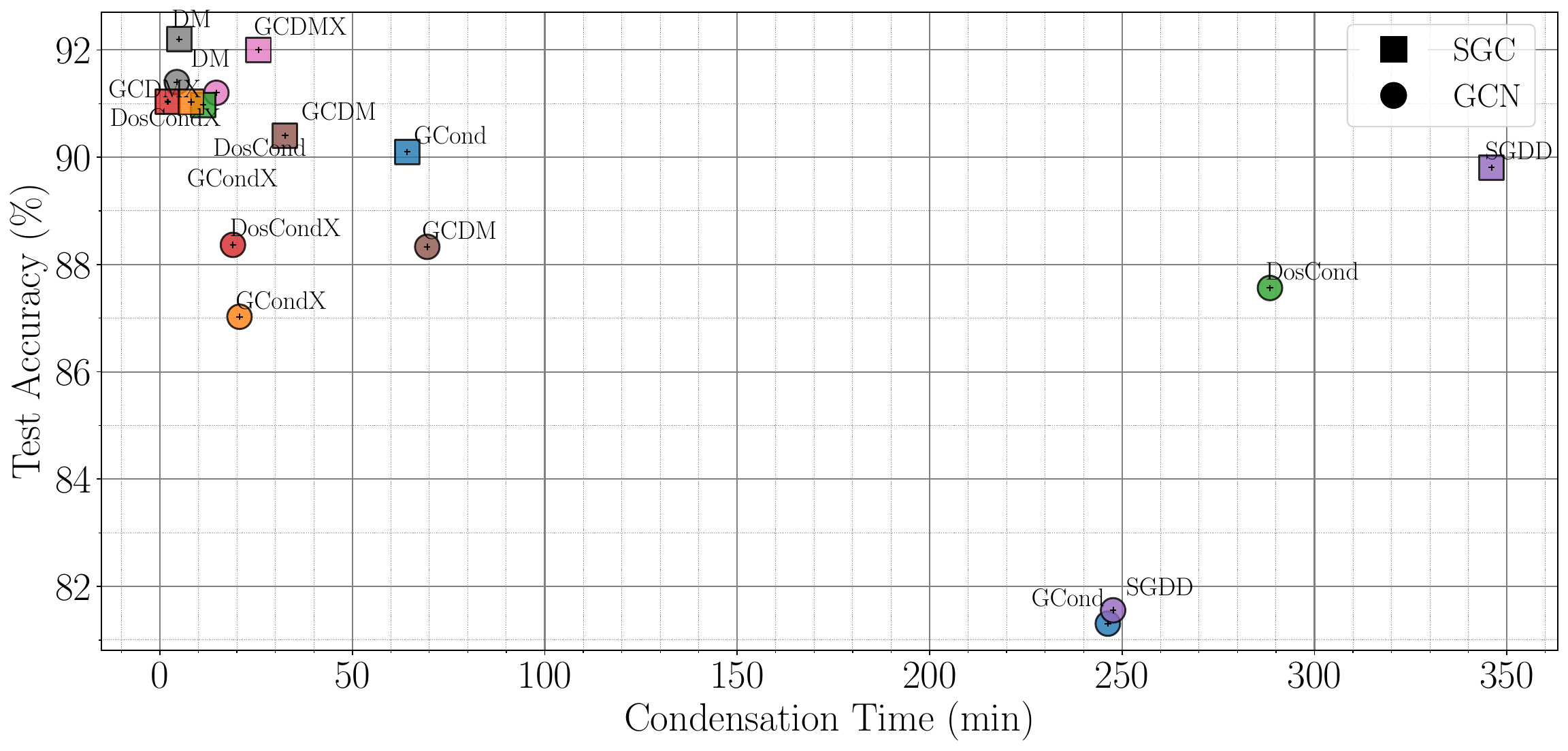}
    \caption{Test accuracy against condensation time for different GC methods on a 153-node condensed graph from the Reddit dataset, with backbone models GCN and SGC.}
    \label{fig:efficiency-reddit}
\end{figure}

In this experiment, more results of the condensation efficiency on all datasets are provided in Figure~\ref{fig:efficiency-citeseer}~$\sim$~\ref{fig:efficiency-reddit}. Overall, distribution matching methods, such as DM, GCDM and GCDMX, can generally achieve a high performance with a high efficiency. While SFGC can sometimes obtain a higher performance, the stability and consistency are not ideal, especially given that SFGC requires an offline generation of expert trajectories.
\begin{figure}[!h]
    \centering
    \includegraphics[width=\linewidth]{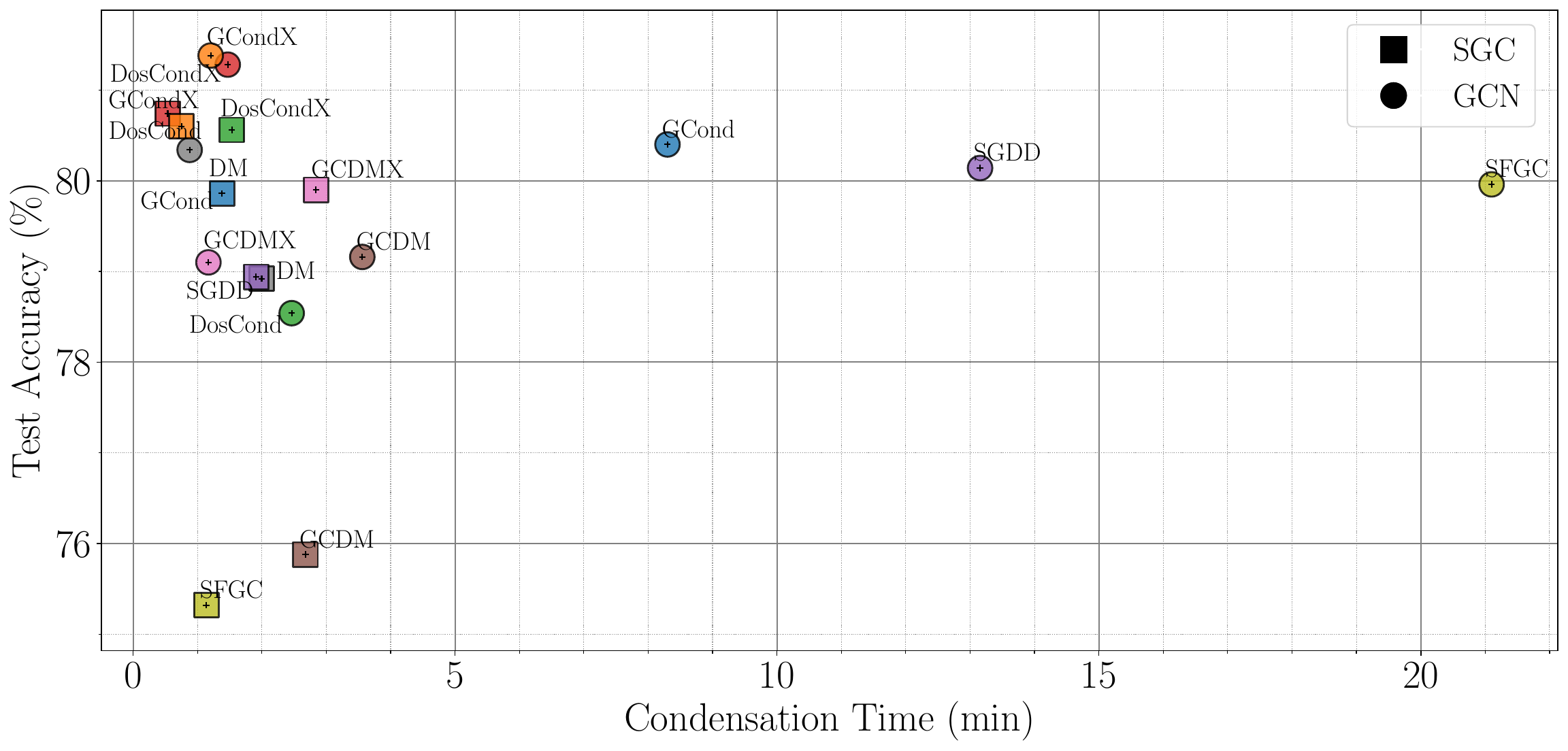}
    \caption{Test accuracy against condensation time for different GC methods on a 35-node condensed graph from the Cora dataset, with backbone models GCN and SGC.}
    \label{fig:efficiency-cora}
\end{figure}
\begin{figure}[!h]
    \centering
    \includegraphics[width=\linewidth]{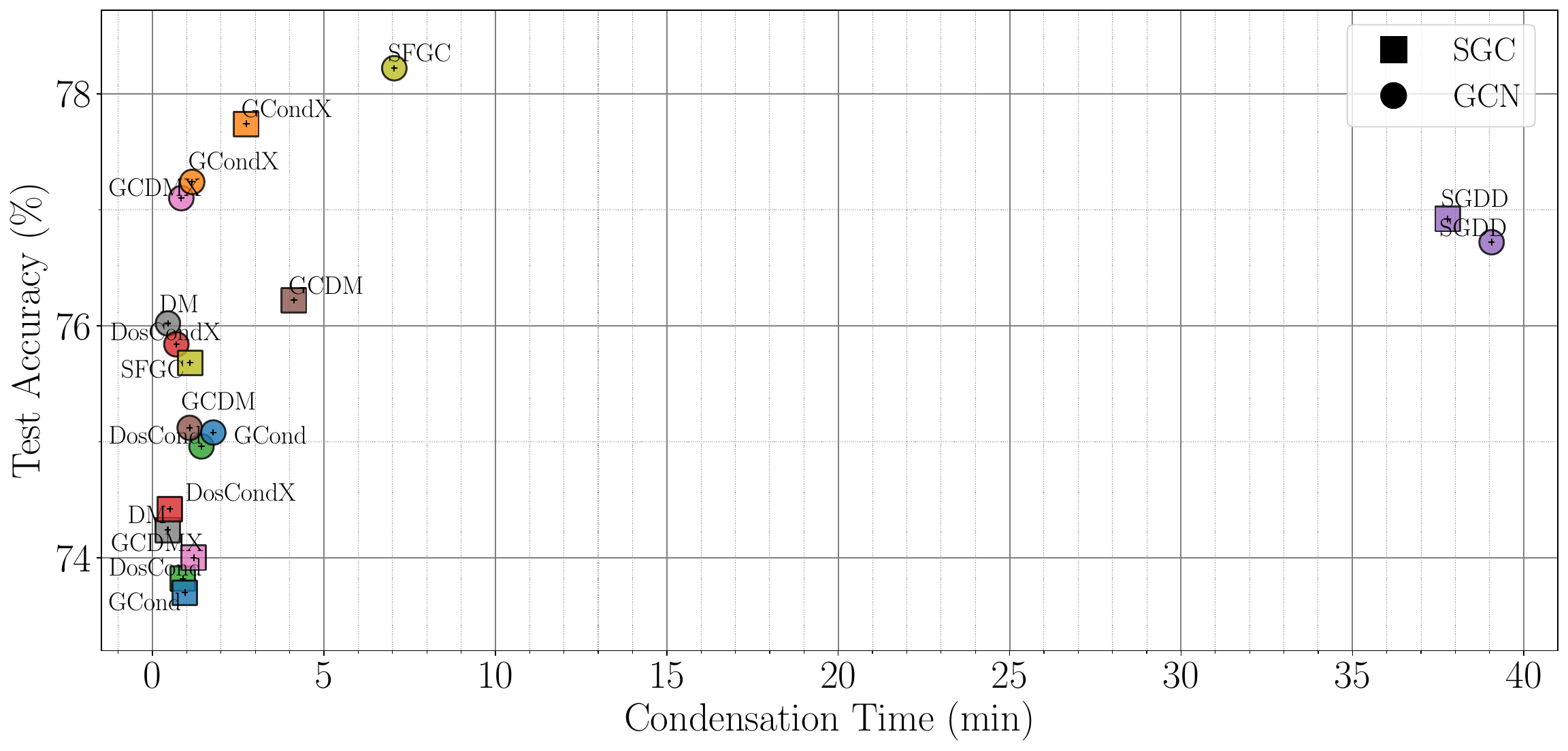}
    \caption{Test accuracy against condensation time for different GC methods on a 15-node condensed graph from the PubMed dataset, with backbone models GCN and SGC.}
    \label{fig:efficiency-pubmed}
\end{figure}
\begin{figure}[!h]
    \centering
    \includegraphics[width=\linewidth]{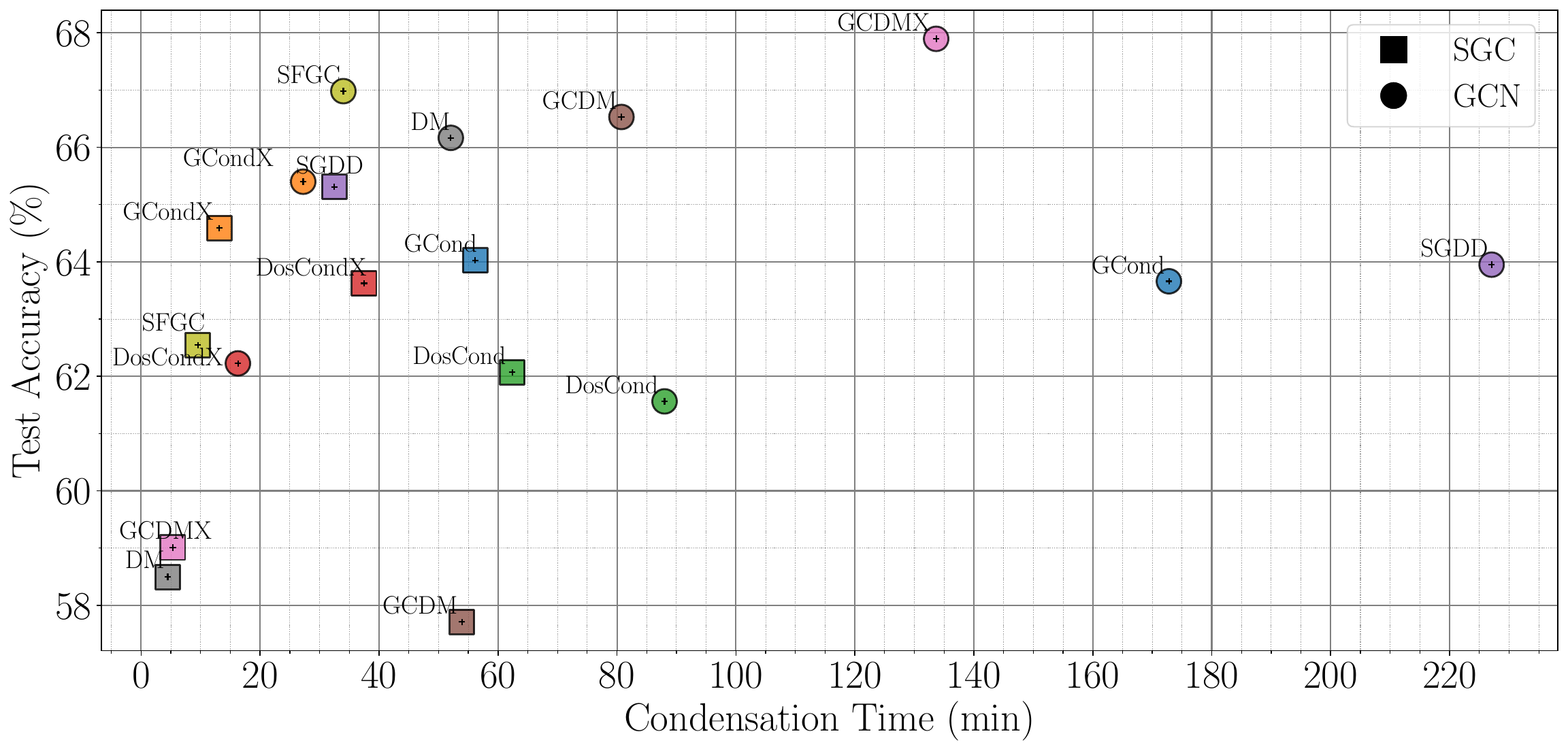}
    \caption{Test accuracy against condensation time for different GC methods on a 612-node condensed graph from the Products dataset, with backbone models GCN and SGC.}
    \label{fig:efficiency-products}
\end{figure}
\begin{figure}[!h]
    \centering
    \includegraphics[width=\linewidth]{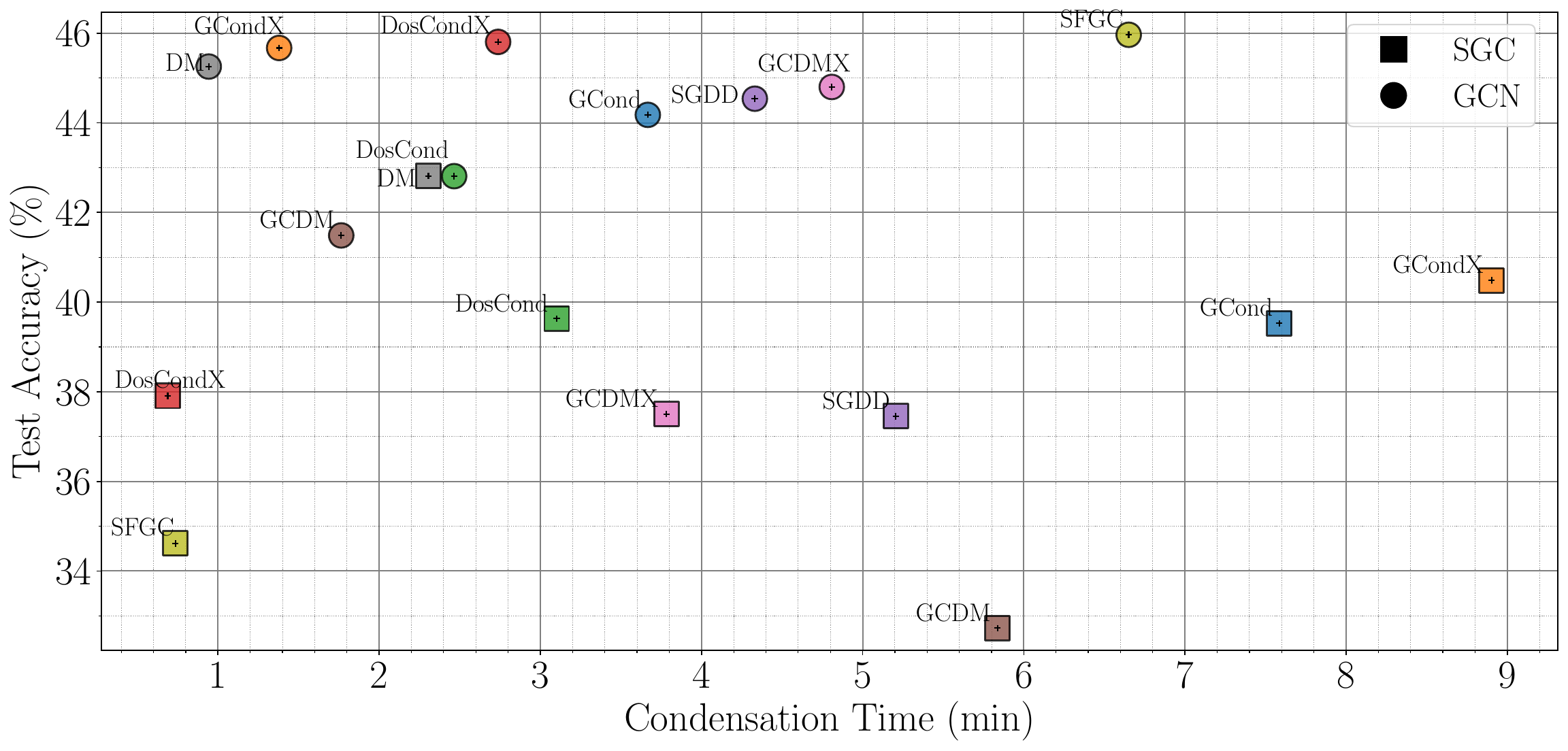}
    \caption{Test accuracy against condensation time for different GC methods on a 44-node condensed graph from the Flickr dataset, with backbone models GCN and SGC.}
    \label{fig:efficiency-flickr}
\end{figure}

\subsection{Cross-architecture Transferability}
\label{sec:exp-transferability-app}
\begin{figure}[!t]
    \centering
    \includegraphics[width=\linewidth]{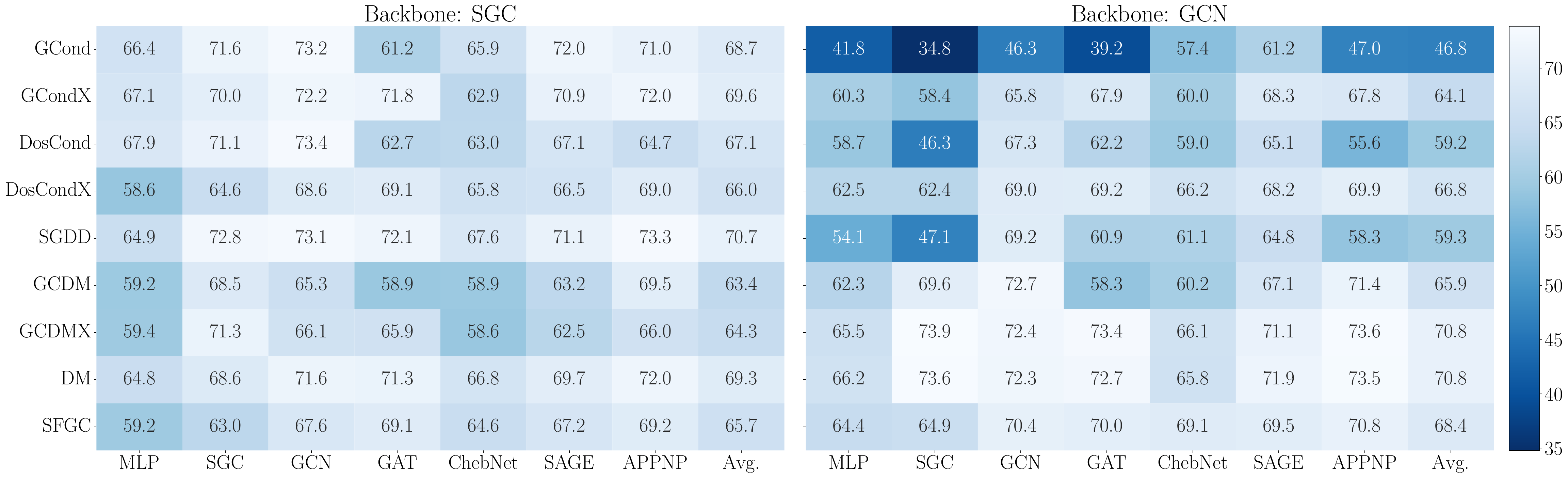}
    \caption{Transferability of condensed graphs for CiteSeer with budget 30.}
    \label{fig:transferability-citeseer}
\end{figure}

\begin{figure}[!t]
    \centering
    \includegraphics[width=\linewidth]{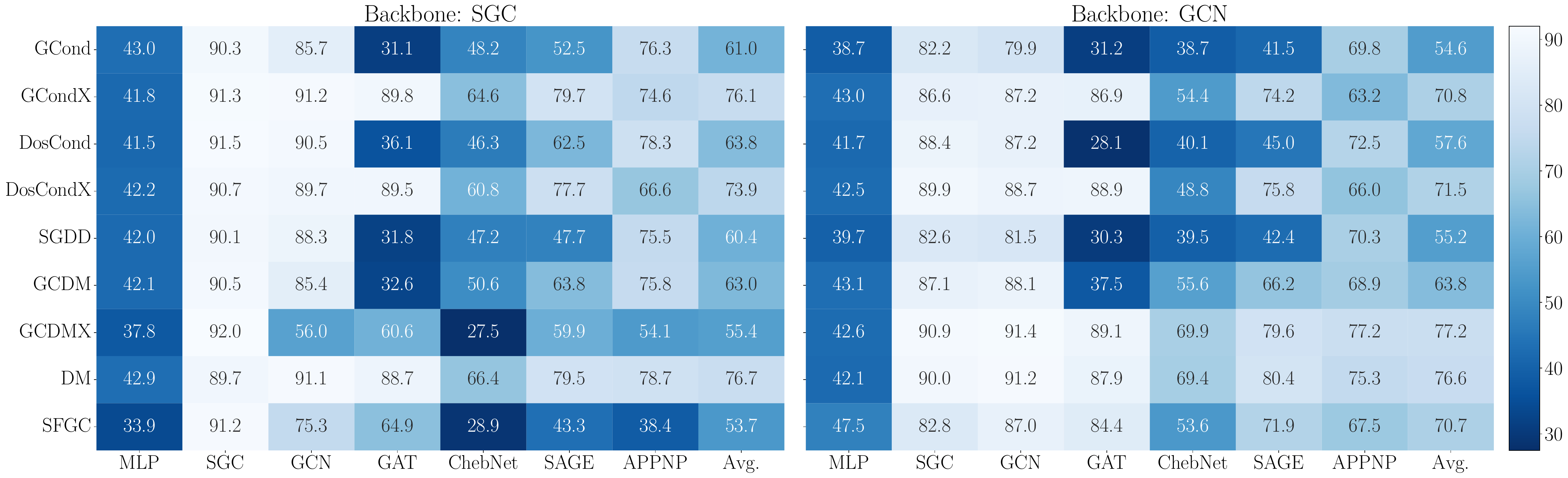}
    \caption{Transferability of condensed graphs for Reddit with budget 153.}
    \label{fig:transferability-reddit}
\end{figure}

In this experiment, more results of the cross-architecture transferability on all datasets are demonstrated in Figure~\ref{fig:transferability-citeseer}~$\sim$~\ref{fig:transferability-reddit}. For small-scale datasets, employing a suitable backbone model generally ensures that all baseline methods achieve comparable cross-architecture performance. However, while a condensed graph may perform well on its original architecture, its performance often deteriorates on alternative architectures when applied to large-scale datasets.

\begin{figure}[!t]
    \centering
    \includegraphics[width=\linewidth]{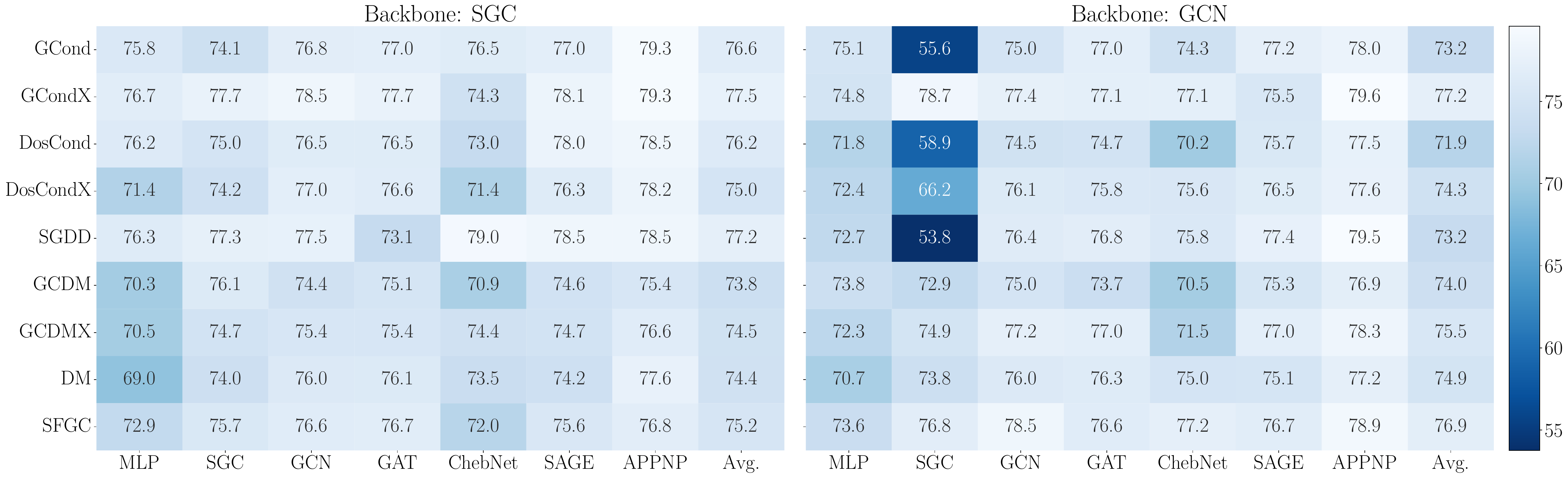}
    \caption{Transferability of condensed graphs for PubMed with budget 15.}
    \label{fig:transferability-pubmed}
\end{figure}
\begin{figure}[!t]
    \centering
    \includegraphics[width=\linewidth]{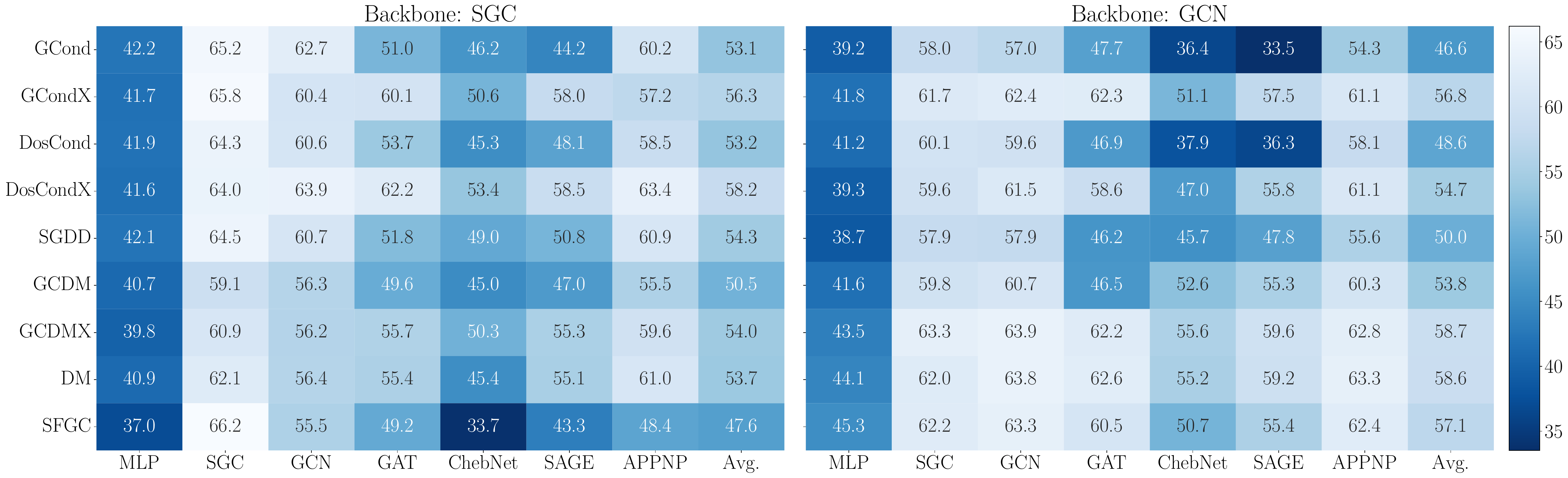}
    \caption{Transferability of condensed graphs for Arxiv with budget 90.}
    \label{fig:transferability-arxiv}
\end{figure}
\begin{figure}[!t]
    \centering
    \includegraphics[width=\linewidth]{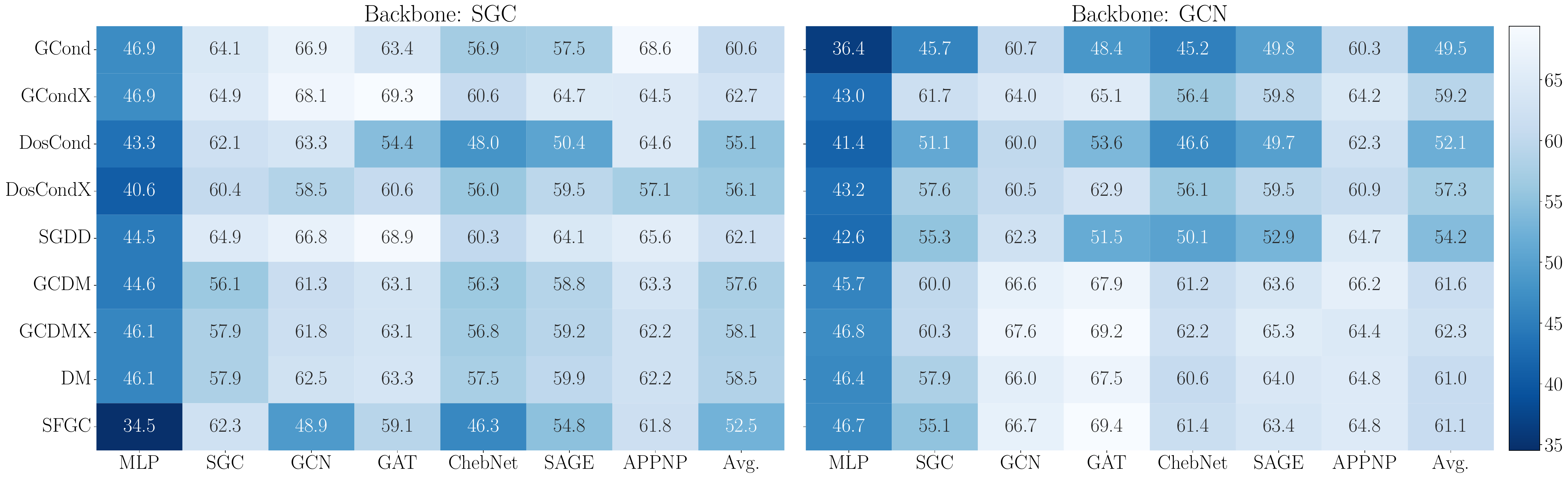}
    \caption{Transferability of condensed graphs for Products with budget 612.}
    \label{fig:transferability-products}
\end{figure}
\begin{figure}[!t]
    \centering
    \includegraphics[width=\linewidth]{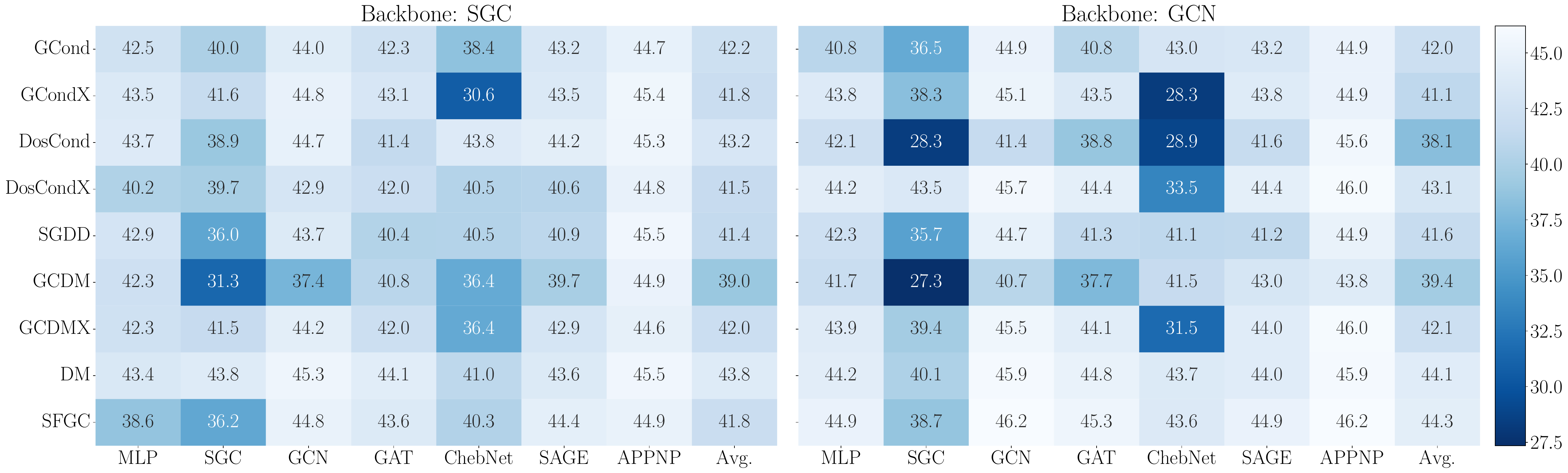}
    \caption{Transferability of condensed graphs for Flickr with budget 44.}
    \label{fig:transferability-flickr}
\end{figure}



\section{Assets License}
Our code base\footnote{https://github.com/superallen13/GCondenser} is open-sourced under the MIT license. All used datasets are public: CiteSeer, Cora, PubMed, Flickr and Reddit can be obtained from DGL~\footnote{https://www.dgl.ai/} or PyG~\footnote{https://www.pyg.org/}; ogbn-arxiv and ogbn-products are provided by OGB~\footnote{https://ogb.stanford.edu/}. 

\section{Ethic Statement}
Our experiments are designed and conducted in strict adherence to ethical standards and do not involve any ethical concerns. We ensure that all procedures are compliant with relevant guidelines and that there is no harm or risk to any participants or environments involved. 